\documentclass[10pt,twocolumn,letterpaper]{article}
\usepackage{cvpr}     
\usepackage{graphicx}
\usepackage{amsmath}
\usepackage{amssymb}
\usepackage{booktabs}
\usepackage{graphbox}
\usepackage{setspace}
\usepackage{makecell}
\usepackage{array}
\usepackage{threeparttable}
\usepackage{multirow}
\usepackage{booktabs}
\usepackage{enumitem}
\usepackage{comment}
\usepackage[pagebackref,breaklinks,colorlinks]{hyperref} 
\usepackage[capitalize]{cleveref}
\crefname{section}{Sec.}{Secs.}
\Crefname{section}{Section}{Sections}
\Crefname{table}{Table}{Tables}
\crefname{table}{Tab.}{Tabs.}

\def\cvprPaperID{5277}  
\def\confName{CVPR}
\def\confYear{2023}
\begin{document}
\title{CiCo: Domain-Aware Sign Language Retrieval via \\Cross-Lingual Contrastive Learning}

\author{
Yiting Cheng$^{1}$, 
\hspace{0.3cm}Fangyun Wei$^{2}$\thanks{Corresponding author.},
\hspace{0.3cm}Jianmin Bao$^{2}$,
\hspace{0.3cm}Dong Chen$^{2}$,
\hspace{0.3cm}Wenqiang Zhang$^{1}\footnotemark[1]$
\\
$^1$School of Computer Science, Fudan University,\hspace{0.5cm} $^2$Microsoft Research Asia \\
{\tt\small \{ytcheng18, wqzhang\}@fudan.edu.cn}, \hspace{0.5cm} {\tt\small \{fawe, jianbao, doch\}@microsoft.com}
}
\newcommand\blfootnote[1]{%
\begingroup
\renewcommand\thefootnote{}\footnote{#1}%
\addtocounter{footnote}{-1}%
\endgroup
}
\maketitle

\begin{abstract}
This work focuses on sign language retrieval—a recently proposed task for sign language understanding. Sign language retrieval consists of two sub-tasks: text-to-sign-video (T2V) retrieval and sign-video-to-text (V2T) retrieval. Different from traditional video-text retrieval, sign language videos, not only contain visual signals but also carry abundant semantic meanings by themselves due to the fact that sign languages are also natural languages. 
Considering this character, we formulate sign language retrieval as a cross-lingual retrieval problem as well as a video-text retrieval task. Concretely, we take into account the linguistic properties of both sign languages and natural languages, and simultaneously identify the fine-grained cross-lingual (i.e., sign-to-word) mappings while contrasting the texts and the sign videos in a joint embedding space. This process is termed as cross-lingual contrastive learning. Another challenge is raised by the data scarcity issue—sign language datasets are orders of magnitude smaller in scale than that of speech recognition. We alleviate this issue by adopting a domain-agnostic sign encoder pre-trained on large-scale sign videos into the target domain via pseudo-labeling. Our framework, termed as domain-aware sign language retrieval via \textbf{C}ross-l\textbf{i}ngual \textbf{Co}ntrastive learning or CiCo for short, outperforms the pioneering method by large margins on various datasets, e.g., +22.4 T2V and +28.0 V2T R@1 improvements on How2Sign dataset, and +13.7 T2V and +17.1 V2T R@1 improvements on PHOENIX-2014T dataset. Code and models are available at: \url{https://github.com/FangyunWei/SLRT}.
\end{abstract}
\section{Introduction}
\label{sec:intro}
Sign languages are the primary means of communication used by people who are deaf or hard of hearing. Sign language understanding~\cite{koller2015continuous,cihan2017subunets,cui2019deep,koller2019weakly,chen2022simple,camgoz2020sign,albanie2020bsl,varol2021read,duarte2022sign,chen2022twostream,zuo2023natural} is significant for overcoming the communication barrier between the hard-of-hearing and non-signers. Sign language recognition and translation (SLRT) has been extensively studied, with the goal of recognizing the \textit{arbitrary} semantic meanings conveyed by sign languages. However, the lack of available data significantly limits the capability of SLRT. In this paper, we focus on developing a framework for a recently proposed sign language retrieval task~\cite{duarte2022sign}. Unlike SLRT, sign language retrieval focuses on retrieving the meanings that signers express from a \textit{closed-set}, which can significantly reduce error rates in realistic deployment.

\begin{figure}
     \centering
     \begin{subfigure}[b]{0.45\textwidth}
         \centering
         \includegraphics[width=\textwidth]{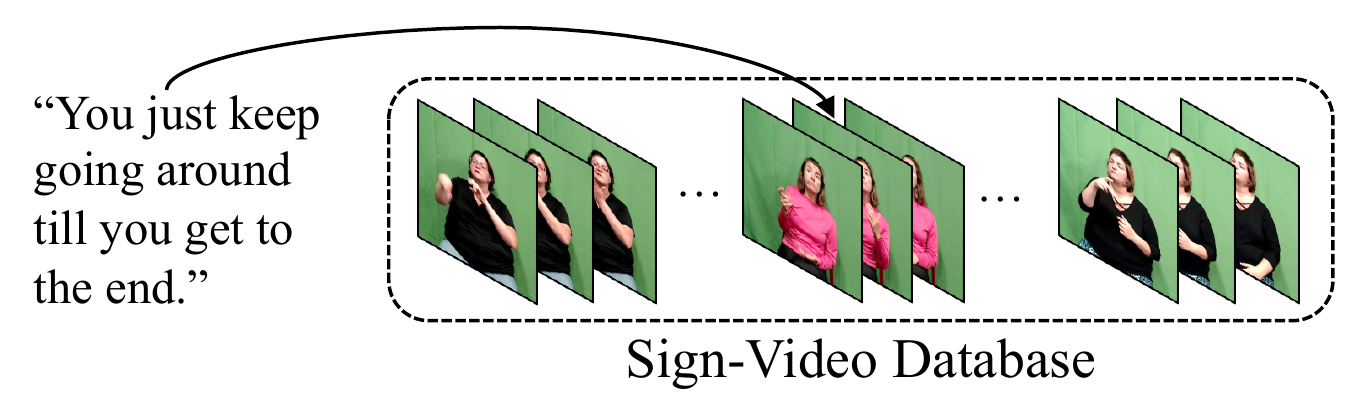}
         \caption{Text-to-sign-video (T2V) retrieval.}
         \label{fig:teaser_A}
     \end{subfigure}
     \hfill
     \begin{subfigure}[b]{0.45\textwidth}
         \centering
         \includegraphics[width=\textwidth]{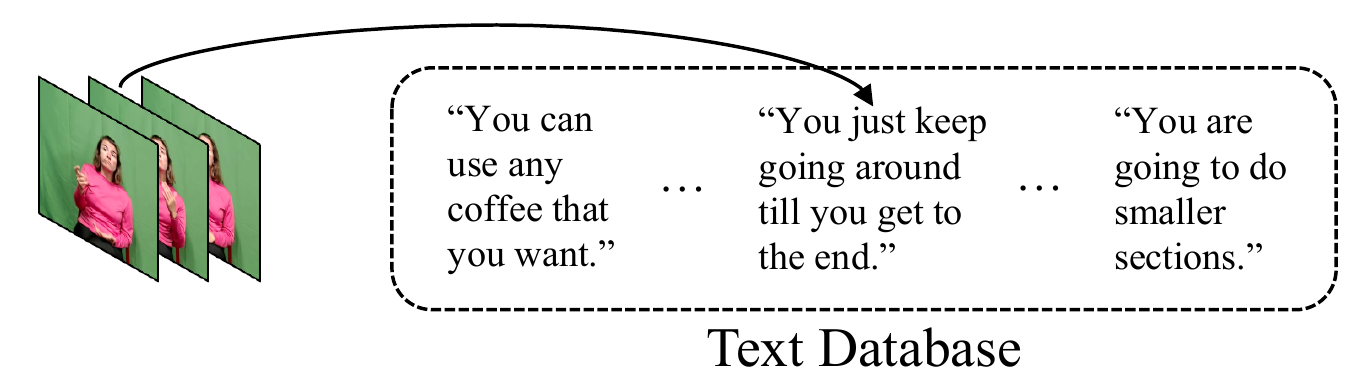}
         \caption{Sign-video-to-text (V2T) retrieval.}
         \label{fig:teaser_B}
     \end{subfigure}
     \hfill
    \caption{Illustration of: (a) T2V retrieval; (b) V2T retrieval.}
    \label{fig:teaser}
    \vspace{-5mm}
\end{figure}

\begin{figure}
     \centering

    \begin{subfigure}[b]{0.45\textwidth}
         \centering
         \includegraphics[width=\textwidth]{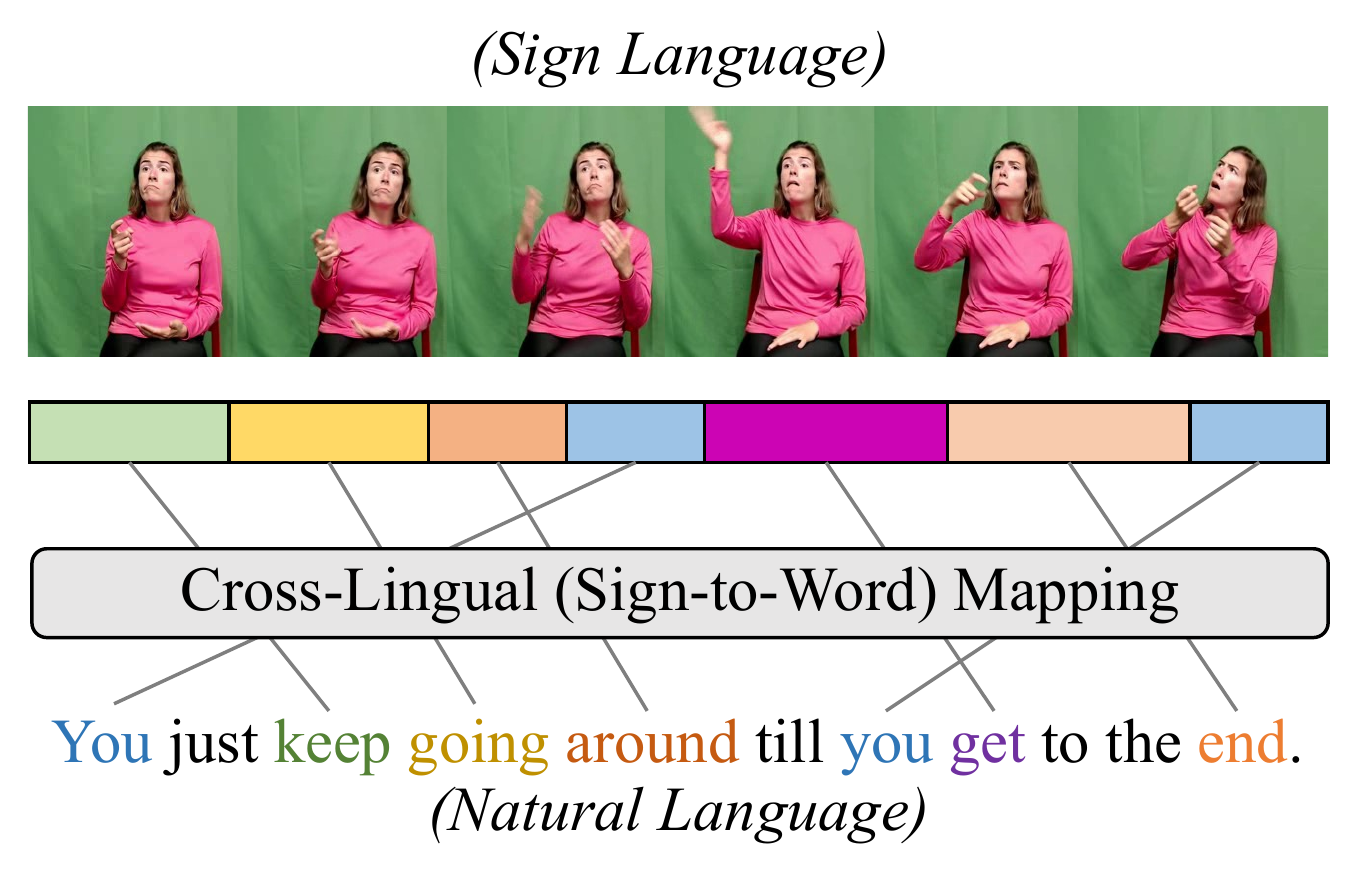}
         \caption{While contrasting the sign videos and the texts in a joint embedding space, we simultaneously identify the fine-grained cross-lingual (sign-to-word) mappings of sign languages and natural languages via the proposed cross-lingual contrastive learning. Existing datasets do not annotate the sign-to-word mappings.}
         \label{fig:teaser_2A}
     \end{subfigure}
     \hfill

     \begin{subfigure}[b]{0.45\textwidth}
         \centering
         \includegraphics[width=\textwidth]{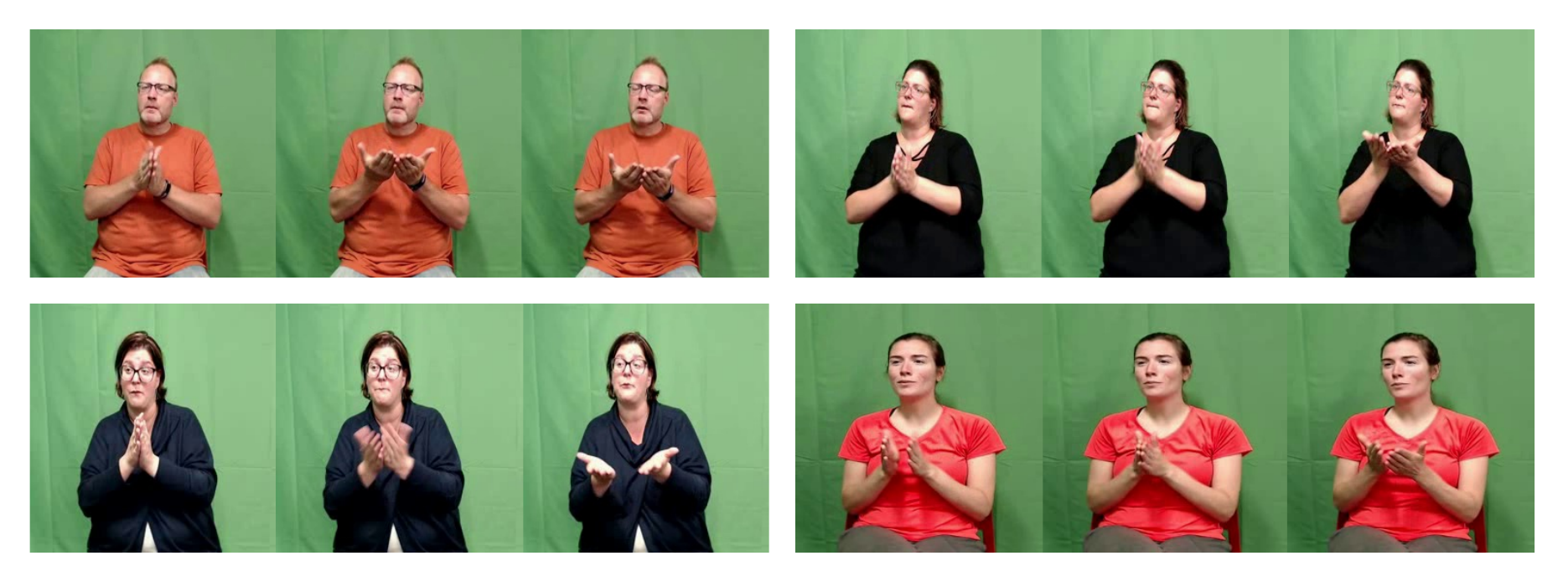}
         \caption{We show four instances of the sign ``Book'' in How2Sign~\cite{Duarte_CVPR2021} dataset, which are identified by our approach. Please refer to the supplementary material for more examples.}
         \label{fig:teaser_2B}
     \end{subfigure}
     \hfill
    \caption{Illustration of: (a) cross-lingual (sign-to-word) mapping; (b) sign-to-word mappings identified by our CiCo.}
    \label{fig:teaser2}
    \vspace{-5mm}
\end{figure}

Sign language retrieval is both similar to and distinct from the traditional video-text retrieval. On the one hand, like video-text retrieval, sign language retrieval is also composed of two sub-tasks, \textit{i.e.}, text-to-sign-video (T2V) retrieval and sign-video-to-text (V2T) retrieval. Given a free-form written query and a large collection of sign language videos, the objective of T2V is to find the video that best matches the written query (Figure~\ref{fig:teaser_A}). In contrast, the goal of V2T is to identify the most relevant text description given a query of sign language video (Figure~\ref{fig:teaser_B}).

On the other hand, different from the video-text retrieval, sign languages, like most natural languages, have their own grammars and linguistic properties. Therefore, sign language videos not only contain visual signals, but also carry semantics (\textit{i.e.}, sign\footnote{We use sign to denote lexical item within a sign language vocabulary.}-to-word mappings between sign languages and natural languages) by themselves, which differentiates them from the general videos that merely contain visual information. Considering the linguistic characteristics of sign languages, we formulate sign language retrieval as a cross-lingual retrieval~\cite{ballesteros1996dictionary,lavrenko2002cross,tran2020cross} problem in addition to a video-text retrieval~\cite{yu2018joint,gabeur2020multi,bain2021frozen,LUO2022293,liu2021hit,liu2019use,sun2019videobert} task.

Sign language retrieval is extremely challenging due to the following reasons: (1) Sign languages are completely separate and distinct from natural languages since they have unique linguistic rules, word formation, and word order. The transcription between sign languages and natural languages is complicated, for instance, the word order is typically not preserved between sign languages and natural languages. It is necessary to automatically identify the sign-to-word mapping from the cross-lingual retrieval perspective; (2) In contrast to the text-video retrieval datasets~\cite{miech2019howto100m,monfort2021spoken} which contain millions of training samples, sign language datasets are orders of magnitude smaller in scale—for example, there are only 30K video-text pairs in How2Sign~\cite{Duarte_CVPR2021} training set; (3) Sign languages convey information through the handshape, facial expression, and body movement, which requires models to distinguish fine-grained gestures and actions; (4) Sign language videos typically contain hundreds of frames. It is necessary to build efficient algorithms to lower the training cost and fit the long videos as well as the intermediate representations into limited GPU memory.   

In this work, we concentrate on resolving the challenges listed above:
\begin{itemize}[leftmargin=*]
    \item We consider the linguistic rules (\textit{e.g.}, word order) of both sign languages and natural languages. We formulate sign language retrieval as a cross-lingual retrieval task as well as a video-text retrieval problem. While contrasting the sign videos and the texts in a joint embedding space as achieved in most vision-language pre-training frameworks~\cite{sun2019videobert,bain2021frozen,LUO2022293}, we simultaneously identify the fine-grained cross-lingual (sign-to-word) mappings between two types of languages via our proposed cross-lingual contrastive learning as shown in Figure~\ref{fig:teaser2}.
    \item Data scarcity typically brings in the over-fitting issue. To alleviate this issue, we adopt transfer learning and adapt a recently released domain-agnostic sign encoder~\cite{varol2021read} pre-trained on large-scale sign-videos to the target domain. Although this encoder is capable of distinguishing the fine-grained signs, direct transferring may be sub-optimal due to the unavoidable domain gap between the pre-training dataset and sign language retrieval datasets. To tackle this problem, we further fine-tune a domain-aware sign encoder on pseudo-labeled data from target datasets. The final sign encoder is composed of the well-optimized domain-aware sign encoder and the powerful domain-agnostic sign encoder. 
    \item In order to effectively model long videos, we decouple our framework into two disjoint parts: (1) a sign encoder which adopts a sliding window on sign-videos to pre-extract their vision features; (2) a cross-lingual contrastive learning module which encodes the extracted vision features and their corresponding texts in a joint embedding space.
\end{itemize}

Our framework, called domain-aware sign language retrieval via \textbf{C}ross-l\textbf{i}ngual \textbf{Co}ntrastive learning or CiCo for short, outperforms the pioneer SPOT-ALIGN~\cite{duarte2022sign} by large margins on various datasets, achieving 56.6 (+22.4) T2V and 51.6 (+28.0) V2T R@1 accuracy (improvement) on How2Sign~\cite{Duarte_CVPR2021} dataset, and 69.5 (+13.7) T2V and 70.2 (+17.1) V2T R@1 accuracy (improvement) on PHOENIX-2014T~\cite{camgoz2018neural} dataset. With its simplicity and strong performance, we hope our approach can serve as a solid baseline for future research.

\section{Related Work}

\noindent\textbf{Sign Language Understanding.} Sign language understanding aims at interpreting the semantic information conveyed within sign videos. Researchers have explored such capability on various tasks including sign language recognition~\cite{fillbrandt2003extraction,farhadi2007transfer,sutton2007mouthings,joze2018MSASL,li2020transferring,chen2022twostream,zuo2023natural}, sign spotting~\cite{albanie2020bsl,momeni20_bsldict,varol2021read,duarte2022sign}, sign language translation~\cite{chen2022simple,camgoz2018neural,zhou2021spatial,camgoz2020multi,li2020tspnet} and our focused sign language retrieval~\cite{duarte2022sign}. 

One of the fundamental tasks of sign language understanding is sign language recognition (SLR), which aims to transcribe a sign video into a gloss sequence. Previous works on SLR focus on designing carefully engineered features~\cite{fillbrandt2003extraction,farhadi2007transfer,sutton2007mouthings} or modeling temporal dependencies~\cite{starner1995visual,starner1998real}. Recently, the success of 3D convolutional neural networks in action related tasks~\cite{wang2018non,lin2019tsm,tran2018closer} is transferred to SLR. In particular, the I3D~\cite{carreira2017quo} architecture has proven to be effective for this task~\cite{joze2018MSASL,li2020transferring,li2020WSASL,albanie2020bsl,varol2021read}. In this work, we also adopt this network architecture in our sign encoder.

Sign spotting is a particular variant of sign language recognition. It aims to localize all instances of a given sign within an untrimmed video. Recent works tackle this task with auxiliary cue of subtitles, introducing automatic annotation systems by using mouthing ~\cite{albanie2020bsl}, dictionaries~\cite{momeni20_bsldict} and attention maps of Transformer~\cite{varol2021read}. SPOT-ALIGN~\cite{duarte2022sign} extends existing spotting methods~\cite{albanie2020bsl,momeni20_bsldict} with an iterative training schema, which alternates between repeated sign spotting and model fine-tuning. In this work, pseudo-labeling is served as our sign spotting approach to localize isolated signs in untrimmed videos from target sign language retrieval datasets. Compared with above efforts, our approach only employs a pre-trained sign encoder without utilizing additional auxiliary cue, which is proved to be simple yet efficient. 

Early works of sign language retrieval primarily investigate query-by-example searching~\cite{athitsos2010large,zhang2010usingRevi}, which queries individual instances with given sign examples. Our work focuses on free-form textual retrieval—a recently introduced task by SPOT-ALIGN~\cite{duarte2022sign}. It symbolizes the real-world scenario of searching sign language videos with natural languages. The pioneer SPOT-ALIGN~\cite{duarte2022sign} purely formulates sign language retrieval as a video-text retrieval task, where the cross-modal alignment is modeled upon overall global embeddings of sign videos and texts. However, the linguistic properties of sign languages are ignored. In contrast, we formulate sign language retrieval as a joint task of text-video retrieval and cross-lingual retrieval.
 
\noindent\textbf{Vision-Language Models.} Learning general-purpose representations for visual and textual modalities is a long-standing topic~\cite{bengio2013representation,lecun2015deep}. The idea has been investigated decades ago~\cite{mori1999image}. Recently, the prominent success of CLIP~\cite{radford2021learning}, ALIGN~\cite{jia2021scaling} and ALBEF~\cite{li2021align} has demonstrated the capability of learning joint cross-modal representations with large-scale web data using simple image-text contrastive learning. Similar idea has also been explored in video-text retrieval, which is the closest area of our work. The common practice in image-text retrieval~\cite{radford2021learning,frome2013devise,gong2014multi,joulin2016learning} and video-text retrieval~\cite{yu2018joint,gabeur2020multi,bain2021frozen,LUO2022293,liu2021hit,liu2019use,sun2019videobert} is to encode the images/videos and texts into the overall representations. It is reasonable since the visual signals of typical image-text and video-text retrieval datasets mainly describe the certain objects or describable events. In contrast, sign videos, as the carriers of sign languages, convey abundant semantics by themselves. There exist fine-grained mappings between sign videos and natural languages. We find that identifying such cross-lingual (sign-to-word) mappings significantly boosts the performance of sign language retrieval.

\noindent\textbf{Cross-Lingual Information Retrieval.} In this work, we also formulate sign-language retrieval as a cross-lingual task. In natural language processing community, cross-lingual information retrieval (CLIR) ~\cite{xu2001evaluating,lavrenko2002cross,tran2020cross} refers to the task of retrieving documents between different languages. One of the most common approaches of CLIR is to learn sentence-level embedding alignment by mapping pre-acquired monolingual embeddings~\cite{mikolov2013exploiting,zhang2019girls,ormazabal2019analyzing} of different languages into a shared space. Recently, researchers have exploited fine-grained word-level mappings with self-training~\cite{artetxe2017learning,artetxe2018robust,tran2020cross}. They find that mappings can be learned by initiating with a seed dictionary and alternating between alignment modeling and dictionary mining. Our work also exploits fine-grained word-level mappings, called sign-to-word mappings, in the context of sign language retrieval. Identifying sign-to-word mappings is challenging since sign languages are expressed in visual modality. The proposed cross-lingual contrastive learning tackles this challenge by exploiting the fine-grained cross-modal interactions.

\section{Methodology}
\begin{figure*}
     \centering
     \begin{subfigure}[b]{0.95\textwidth}
         \centering
         \includegraphics[width=\textwidth]{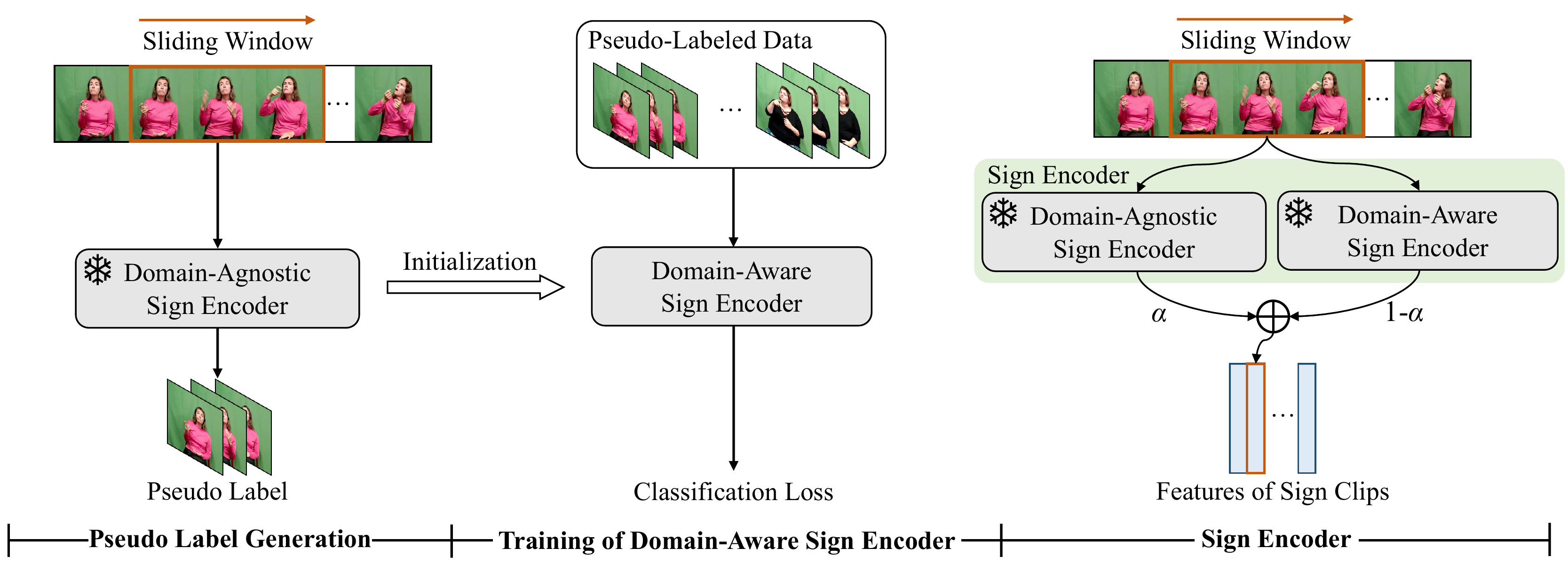}
         \caption{Illustration of sign encoder.}
         \label{fig:overview_A}
     \end{subfigure}
     \hfill
     \begin{subfigure}[b]{0.95\textwidth}
         \centering
         \includegraphics[width=\textwidth]{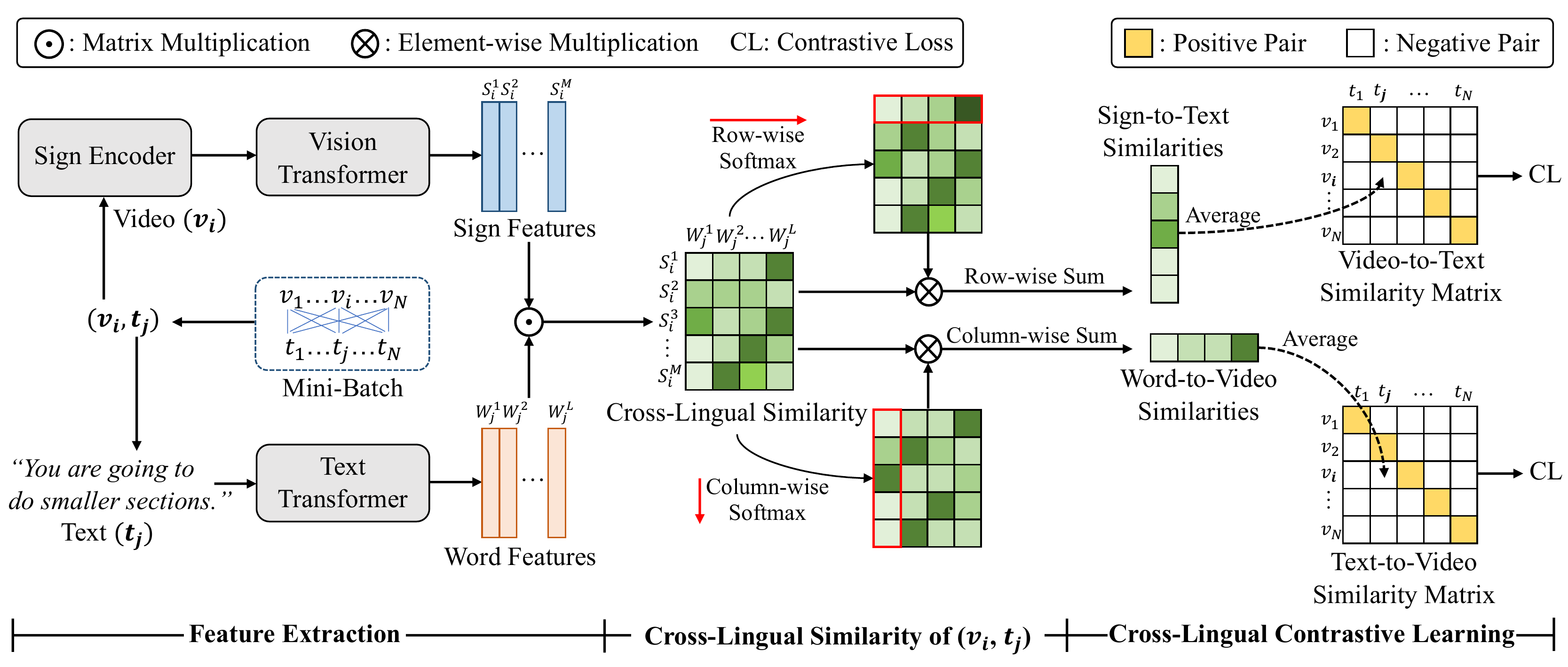}
         \caption{Illustration of cross-lingual contrastive learning.}
         \label{fig:overview_B}
     \end{subfigure}
     \hfill
    \caption{Overview of our framework. (a) Sign encoder is composed of a powerful domain-agnostic sign encoder pre-trained on large-scale sign videos, and a domain-aware sign encoder fine-tuned on pseudo-labeled data from target datasets. We adopt a sliding window manner to extract a discriminative and domain-aligned feature per clip. (b) Cross-lingual contrastive learning takes $N$ sign-video-text pairs as inputs and contrasts paired data in a shared embedding space while implicitly identifying the fine-grained sign-to-word mappings during training.
    %Sign feature domain alignment (SFDA) module aims to provide a strong sign feature for subsequent cross-modal learning. Without sign-level annotation, SFDA turn to generalize a powerful domain-agnostic sign encoder to target dataset with pseudo labeling. The fine-tuned domain-aware sign encoder is expected to better align in the target sign language domain. Besides, a feature fusion strategy is introduced to make full use of two complementary sign encoders. (b) Cross-lingual contrastive learning (CLCL) aims to handle the retrieval task by learning a joint embedding space for sign video features and texts. Fine-grained cross-modal interactions are proposed in CLCL to model the complicated sign-to-word mappings between sign language and natural language.
    }
    \label{fig:overview}
        \vspace{-5mm}
\end{figure*}

In this section, we first formulate the task of sign language retrieval in Section~\ref{sec:task}. Next, we introduce our framework which is composed of two parts: 1) a sign encoder which extracts discriminative and domain-aligned features of sign videos (Section~\ref{Subsec:SFDA}); 2) a cross-lingual contrastive learning framework, which contrasts sign-video-text pairs while concurrently identifying the fine-grained sign-to-word mappings (Section~\ref{Subsec:SWCL}). At last, we explore text augmentations in Section~\ref{Subsec:TA}.

\subsection{Task Formulation}
\label{sec:task}
Let $\mathcal{V}$ and $\mathcal{T}$ denote a set of sign videos and their corresponding texts (transcriptions), respectively. Sign language retrieval consists of two tasks namely text-to-sign-video retrieval (T2V), and sign-video-to-text retrieval (V2T), respectively. The objective of T2V is to find the sign video $v\in\mathcal{V}$ whose signing content best matches the text query. In contrast, the reverse task V2T requires the model to identify the most relevant text (transcription) $t\in\mathcal{T}$ given a query of sign video. We resolve sign language retrieval by learning a joint embedding space of sign videos and texts.

\subsection{Sign Encoder}
\label{Subsec:SFDA}
\noindent\textbf{Process Sign Videos with Sliding Window.} Sign videos from sign language retrieval datasets typically contain hundreds of frames. To efficiently train our model and lower the usage of GPU memory, given a sign video, we adopt a sliding window manner with stride of 1 and window size of 16 to produce $M$ temporally overlapped clips. Next, we separately feed each clip into a sign encoder to extract its feature. The final sign-video feature is yielded by stacking features coming out of $M$ clips along temporal dimension. A powerful sign encoder is crucial.

\noindent\textbf{Overview of Sign Encoder.} Recent advances in sign spotting~\cite{momeni20_bsldict,varol2021read} greatly facilitate the collection of large-scale sign language datasets, enabling powerful representation learning abilities of convolutional neural networks on the sign classification task. Previous methods~\cite{duarte2022sign,das2022machine} have demonstrated the feasibility of transferring a sign encoder pre-trained on large-scale sign-spotting data into downstream tasks. We follow this practice and use an I3D network pre-trained on BSL-1K~\cite{varol2021read}, a sign classification dataset collected via sign spotting, as our primary sign encoder. Due to its favorable transfer performance, we term this model as a domain-agnostic sign encoder. Nevertheless, the domain gap between BSL-1K and sign language retrieval datasets is non-negligible. To tackle this problem, we further fine-tune a domain-aware sign encoder, which has an identical architecture to the domain-agnostic sign encoder, on target datasets through pseudo-labeling. The final sign encoder is composed of the well-optimized domain-aware sign encoder and the powerful domain-agnostic sign encoder, as illustrated in Figure~\ref{fig:overview_A}.

\noindent\textbf{Pseudo-Labeling on Target Datasets.} Now we describe the details of pseudo-labeling. Given a sign video from a target dataset, we adopt a sliding window with the stride of 1 and the window size of 16 to generate a set of temporally overlapped clips. For each clip, we first utilize the pre-trained domain-agnostic sign encoder to produce its prediction. Then we binarize the prediction with a pre-defined threshold $\lambda$ to generate the corresponding pseudo label. The invalid samples, whose maximum score is lower than $\lambda$, are filtered. We repeat the above process for all sign videos and eventually build a pseudo-labeled set. Our domain-aware sign encoder, which is initialized by the domain-agnostic sign encoder, is fine-tuned on the pseudo-labeled set via a standard cross-entropy loss for classification training.

\noindent\textbf{Feature Extraction with Sign Encoder.}
So far, we acquire a domain-aware sign encoder approximately aligned in target domain.
Nevertheless, its capability is restricted by the unavoidable noises in pseudo-labels and the limited amount of pseudo-labeled samples.
Recall that we already have a powerful domain-agnostic sign encoder pre-trained on large-scale dataset in hand, inspiring us to make use of both domain-agnostic sign encoder $h_{\xi}(\cdot)$ and domain-aware sign encoder $h_{\theta}(\cdot)$ to extract discriminative and domain-aligned features. Our final sign encoder $H(\cdot)$, as shown in Figure~\ref{fig:overview_A}, is a weighted combination of $h_{\xi}(\cdot)$ and $h_{\theta}(\cdot)$ with a trade-off hyper-parameter $\alpha$. As described above, $H(\cdot)$ encodes sign videos in a sliding window manner. For simplicity, we use $H(v)$ to denote feature extraction on sign video $v$, which is formulated as: 
\begin{equation}
 \label{eq:feature_fusion}
 H(v) = \alpha h_{\xi}(v) + (1-\alpha) h_{\theta}(v).
\end{equation}

\subsection{Cross-Lingual Contrastive Learning}
\label{Subsec:SWCL}
The objective of cross-lingual contrastive learning (CLCL) is to learn a joint embedding space of sign videos and texts while concurrently identifying the fine-grained sign-to-word mappings during training. An overview is shown in Figure~\ref{fig:overview_B}. CLCL takes a mini-batch $\{(v_{n}, t_{n})\}^{N}_{
n=1}$ containing $N$ sign-video-text pairs as input, and contrasts paired data in a shared embedding space for sign language retrieval.

\noindent\textbf{Sign Features and Word Features.}
Given a sign video $v \in \{v_{n}\}^{N}_{
n=1}$, we first adopt our sign encoder $H(\cdot)$ described in Section~\ref{Subsec:SFDA} to extract its intermediate feature. Note $H(\cdot)$ encodes sign videos in a sliding window manner, and thus there are no interactions among different clips. To facilitate information exchange, we further append a 12-layer Transformer~\cite{Attention17} $F(\cdot)$ onto $H(\cdot)$ to extract sign features $\boldsymbol{S}$ of sign video $v$, which is formulated as $\boldsymbol{S}=F(H(v)) \in 
\mathbb{R}^{M \times D}$, where $M$ denotes the number of clips, and $D$ is the hidden dimension. Given a text $t \in \{t_{n}\}^{N}_{n=1}$, we convert $t$ into a lower-cased byte pair encoding (BPE) representation~\cite{sennrich2015neural}, which is subsequently fed into another 12-layer Transformer $G(\cdot)$ to generate the word features $\boldsymbol{W} = G(t) \in \mathbb{R}^{L \times D}$, where $L$ represents word number. 

Since CLIP~\cite{radford2021learning} shows excellent transfer capability in various downstream tasks~\cite{fort2021exploring,shen2021much,du2022learning,xu2022simple,xu2023side,huang2022unsupervised}, we initialize $F(\cdot)$ and $G(\cdot)$ with CLIP's image encoder (ViT-B) and text encoder, respectively, to ease the learning. Though CLIP's vision encoder takes image patches as inputs, we experimentally find that it generalizes well in our scenario where input data is in a different modality. 

\noindent\textbf{Cross-Lingual Similarity.} 
There exist inherent sign-to-word mappings between sign languages and natural languages. To incorporate this prior knowledge into learning, we introduce cross-lingual similarity—an indicator to identify sign-to-word mappings between $i$-th sign video $v_i$ and $j$-th text $t_j$. Concretely, given sign features $\boldsymbol{S}_i \in \mathbb{R}^{M \times D}$ of $v_i$ , and word features $\boldsymbol{W}_j \in \mathbb{R}^{L \times D}$ of $t_j$, we calculate a cross-lingual similarity matrix $\boldsymbol{E}_{(i,j)} =\boldsymbol{S}_i \cdot \boldsymbol{W}_j^T \in \mathbb{R}^{M \times L}$. Each element in $\boldsymbol{E}_{(i,j)}$ represents the similarity of a sign clip in $v_i$ and a word in $t_j$.

\noindent\textbf{Cross-Lingual Contrastive Learning.} Directly apply supervisions on token-wise similarity matrix $\boldsymbol{E}_{(i,j)}$ is infeasible due to the absence of fine-grained sign-to-word annotations. Inspired by the recent progress of vision-language contrastive learning~\cite{radford2021learning,li2021align,li2021supervision,frome2013devise,gong2014multi,joulin2016learning}, we turn to contrast the global representations of sign videos and texts. The underlying idea is to calculate a global similarity $z$ of $v_i$ and $t_j$ based on $\boldsymbol{E}_{(i,j)} \in \mathbb{R}^{M \times L}$. 

\noindent\textbf{Sign-Video-to-Text Contrast.} We first introduce sign-video-to-text contrast as shown in Figure~\ref{fig:overview_B}. To be specific, we utilize a Softmax operation to each row of $\boldsymbol{E}_{(i,j)}$, and multiply the resulting matrix with $\boldsymbol{E}_{(i,j)}$ to generate a sign-to-word similarity matrix $\boldsymbol{E}'_{(i,j)} \in \mathbb{R}^{M \times L}$, where each row represents the re-weighted similarities between a sign clip in $v_i$ and all words in $t_j$. After that, we adopt a row-wise addition operation on $\boldsymbol{E}'_{(i,j)}$ to yield the sign-to-text similarity vector $\boldsymbol{e}_{(i,j)} \in \mathbb{R}^{M}$, where each element denotes a similarity of a sign clip in $v_i$ and whole text $t_j$. At last, we average all elements in $\boldsymbol{e}_{(i,j)}$ to produce the global similarity $z$ of sign video $v_i$ and text $t_j$.

In the same way, we can calculate the similarities for both positive pairs $\{(v_{i}, t_{i})\}^{N}_{
i=1}$ and negative pairs $\{(v_{i}, t_{j})\}^{N}_{
i=1,j=1,i\neq j}$ in a mini-batch, yielding a video-to-text similarity matrix $\boldsymbol{Z}_{V2T} \in \mathbb{R}^{N \times N}$, where $\boldsymbol{Z}^{(i,j)}_{V2T}$ denotes the global similarity of $v_i$ and $t_j$. Following CLIP~\cite{radford2021learning}, we adopt InfoNCE loss~\cite{gutmann2010noise} to pull the embeddings of matched image-text pairs together while pushing those of non-matched pairs apart, which is formulated as follows:
\begin{equation}
  \label{loss:NCE}
  \begin{split}
\mathcal{L}_{V2T}= & -\frac{1}{2N}\sum_{i=1}^{N} log \frac{exp(\boldsymbol{Z}_{V2T}^{(i,j)}/\tau)}{\sum_{j=1}^{N} exp(\boldsymbol{Z}_{V2T}^{(i,j)}/\tau)} \\
&-\frac{1}{2N}\sum_{j=1}^{N} log \frac{exp(\boldsymbol{Z}_{V2T}^{(i,j)}/\tau)}{\sum_{i=1}^{N} exp(\boldsymbol{Z}_{V2T}^{(i,j)}/\tau)}, 
  \end{split}
\end{equation}
where $\tau$ is a trainable temperature parameter. 

\noindent\textbf{Text-to-Sign-Video Contrast.}
Up to now, we have introduced sign-video-to-text contrast. A symmetrical version, termed text-to-sign-video contrast, shares the similar spirit as shown in Figure~\ref{fig:overview_B}. The implementation of text-to-sign-video contrast is extremely simple: we replace the row-wise operations (\textit{i.e.}, Softmax and addition) in sign-video-to-text contrast with column-wise ones and keep the remaining processes unchanged. We use $\mathcal{L}_{T2V}$ to denote the loss function of text-to-sign-video contrast. In our implementation, we reuse the loss defined in Eq~\ref{loss:NCE} but substitute the input with the text-to-video similarity matrix $\boldsymbol{Z}_{T2V}$.

\noindent\textbf{Loss Function.}
The overall loss for cross-lingual contrastive learning is a weighted sum of $\mathcal{L}_{V2T}$ and $\mathcal{L}_{T2V}$ with a trade-off hyper-parameter $\beta$:
\begin{equation}
\label{eq:loss:overall}
    \mathcal{L} = \beta \mathcal{L}_{V2T} + (1-\beta) \mathcal{L}_{T2V}.
\end{equation}

\subsection{Text Augmentation}
\label{Subsec:TA}
Considering that the datasets of sign language retrieval are typically small-scale, we explore text augmentations to improve the generalization of our approach. EDA~\cite{eda2019} introduces three simple yet efficient data augmentations in text classification task: random delete randomly removes words in a sentence; synonym replacement randomly selects words from a sentence that are not stop words and replaces them with synonyms; random swap randomly chooses two words in a sentence and swaps their positions. The first two augmentations have been proven effective in text classification task. However, we experimentally find that our focused sign language retrieval is sensitive to random delete and synonym replacement augmentations. To guarantee that the augmented texts preserve the original semantic meanings, we only adopt the random swap augmentation in our approach. We suppose there are two reasons: 1) the word order of sign languages and natural languages are constitutionally distinct, and reordering does not affect semantic meanings; 2) the proposed cross-lingual contrastive learning is insensitive to word order.

\section{Experiment \label{Sec:Exp}}

\subsection{Datasets and Implementation Details}
\noindent\textbf{Datasets.} We primarily focus on \textit{How2Sign}~\cite{Duarte_CVPR2021} dataset. Our model is also evaluated on \textit{PHOENIX-2014T}~\cite{camgoz2018neural} and \textit{CSL-Daily}~\cite{zhou2021improving}, which are primarily used for sign language recognition and translation in previous works.

\textit{How2Sign} is a large-scale continuous American sign language (ASL) dataset consisting of a parallel corpus of about 80 hours of sign videos with subtitle annotations. It covers a wide range of instructional videos corresponding to various categories. There are 31164, 1740 and 2356 sign-video-text pairs in training, validation and test sets, respectively. Following SPOT-ALIGN~\cite{duarte2022sign}, we remove the invalid pairs where the subtitle alignment is detected to exceed the video duration, remaining 31085, 1739 and 2348 available pairs in training, validation and test sets, respectively. The resolution of sign videos is 1280$\times$720, we crop the human bodies of signers with Faster R-CNN~\cite{ren2015faster} to generate valid videos.

\textit{PHOENIX-2014T} is a German sign language (Deutsche Gebärdensprache, DGS) dataset collected in the domain of weather forecast from TV broadcast, consisting of 7096, 519 and 642 video-text pairs in training, validation and test sets, respectively.

\textit{CSL-Daily} is a recently released Chinese sign language (CSL) dataset. The topic of CSL-Daily revolves around people's daily lives, including 18401, 1077, 1176 parallel samples in training, validation and test set, respectively. 

\noindent\textbf{Evaluation Metric.} Following previous works~\cite{duarte2022sign,radford2021learning,liu2019use,sun2019videobert}, retrieval performance is evaluated by recall at rank K (R@K, higher is better) and median rank (MedR, lower is better). We evaluate our approach on both text-to-sign-video (T2V) retrieval and sign-video-to-text (V2T) retrieval tasks. We report R@1, R@5, and R@10 in all experiments, and additionally report MedR when comparing with state-of-the-art approaches. 

\noindent\textbf{Implementation Details.}
The sign encoder takes videos of resolution of 256$\times$256 as input. The domain-agnostic sign encoder is a I3D~\cite{carreira2017quo} network pre-trained on BSL-1K~\cite{varol2021read}. In pseudo label generation, we set the threshold $\lambda$ to 0.6 to filter samples with low-confidence. Non-maximum suppression (NMS) with a temporal window of 24 frames is utilized to remove the duplicates among the pseudo-labeled samples. A collection of approximate 64K pseudo-labeled samples covering a vocabulary of 1220 words is eventually generated. The domain-aware sign encoder is initialized with the domain-agnostic one and fine-tuned with a learning rate of $1\times 10^{-2}$ and batch size of 4 for 15 epochs. In the training of cross-lingual contrastive learning, the vision transformer and text transformer are initialized by the image encoder and text encoder in CLIP (ViT-B/32)~\cite{radford2021learning}. The maximum length of sign clip features and text features are set to 64 and 32, respectively. The model is fine-tuned with Adam optimizer~\cite{KingmaB14} with batch size of 512. The initial learning rate is set to $1\times 10^{-5}$, which is decreased with a cosine schedule following the CLIP~\cite{radford2021learning}. We set $\alpha=0.8$ in Eq.~\ref{eq:feature_fusion} and $\beta=0.5$ in Eq.\ref{eq:loss:overall}. The languages of \textit{PHOENIX-2014T} and \textit{CSL-Daily} are German and Chinese respectively. Since CLIP is trained on English corpus, to reuse CLIP's text encoder, we utilize Google translation to translate the texts of these two datasets into English.

\subsection{Comparison with State-of-the-art Methods}

 \begin{table}[t!]
 \scriptsize
 \centering
 \setlength{\tabcolsep}{1pt}
\begin{tabular}{c|cccc|cccc}
\toprule
\multirow{2}{*}{Model} & \multicolumn{4}{c|}{T2V} & \multicolumn{4}{c}{V2T} \\
 & R@1$\uparrow$  & R@5$\uparrow$ & R@10$\uparrow$ & MedR$\downarrow$ & R@1$\uparrow$  & R@5$\uparrow$ & R@10$\uparrow$ & MedR$\downarrow$ \\
\midrule
SA-SR~\cite{duarte2022sign} & 18.9 & 32.1 & 36.5 & 62.0 & 11.6 & 27.4 & 32.5 & 69.0 \\
SA-CM~\cite{duarte2022sign} & 24.3 & 40.7 & 46.5 & 16.0 & 17.9 & 40.1 & 46.9 & 14.0 \\
SA-COMB~\cite{duarte2022sign} & 34.2 & 48.0 & 52.6 & 8.0 & 23.6 & 47.0 & 53.0 & 7.5 \\
\midrule
\textbf{Ours} & \textbf{56.6} & \textbf{69.9} & \textbf{74.7} & \textbf{1.0} & \textbf{51.6} & \textbf{64.8} & \textbf{70.1} & \textbf{1.0}\\
\bottomrule
\end{tabular}
\caption{Comparison with the different variants of the pioneer SPOT-ALIGN (SA)~\cite{duarte2022sign} on How2Sign~\cite{Duarte_CVPR2021} dataset.}
\vspace{-1em}
\label{Tab:comparison_SOTA_how2sign}
\end{table}

\begin{table}[t!]
 \scriptsize
 \centering
 \setlength{\tabcolsep}{1pt}
\begin{tabular}{c|cccc|cccc}
\toprule
\multirow{2}{*}{Model} & \multicolumn{4}{c|}{T2V} & \multicolumn{4}{c}{V2T} \\
 & R@1$\uparrow$  & R@5$\uparrow$ & R@10$\uparrow$ & MedR$\downarrow$ & R@1$\uparrow$  & R@5$\uparrow$ & R@10$\uparrow$ & MedR$\downarrow$ \\
 \midrule
Translation~\cite{camgoz2020sign}  & 30.2 & 53.1 & 63.4 & 4.5 & 28.8 & 52.0 & 60.8 & 56.1 \\
SA-CM~\cite{duarte2022sign}  & 48.6 & 76.5 & 84.6 & 2.0 & 50.3 & 78.4 & 84.4 & \textbf{1.0} \\
SA-COMB~\cite{duarte2022sign}  & 55.8 & 79.6 & 87.2 & \textbf{1.0} & 53.1 & 79.4 & 86.1 & \textbf{1.0} \\
 \midrule
\textbf{Ours}  & \bf69.5	&\bf86.6 &	\bf92.1	&\bf1.0 & \bf70.2& \bf88.0&  \bf92.8& \bf1.0\\
\bottomrule
\end{tabular}
\caption{Comparison with the different variants of the pioneer SPOT-ALIGN (SA)~\cite{duarte2022sign} on PHOENIX2014T~\cite{camgoz2018neural} dataset.}
\label{Tab:comparison_SOTA_PHOENIX}
\vspace{-1em}
\end{table}

\begin{table}[t!]
 \scriptsize
 \centering
 \setlength{\tabcolsep}{1pt}
\begin{tabular}{c|cccc|cccc}
\toprule
\multirow{2}{*}{Model} & \multicolumn{4}{c|}{T2V} & \multicolumn{4}{c}{V2T} \\
 & R@1$\uparrow$  & R@5$\uparrow$ & R@10$\uparrow$ & MedR$\downarrow$ & R@1$\uparrow$  & R@5$\uparrow$ & R@10$\uparrow$ & MedR$\downarrow$ \\
 \midrule
Ours &75.3 & 88.2 & 91.9 &1.0&74.7&89.4&92.2&1.0\\
\bottomrule
\end{tabular}
\caption{We additionally provide a baseline for CSL-Daily~\cite{zhou2021improving} dataset.}
\label{Tab:baseline_CSL}
\vspace{-6mm}
\end{table}

We compare our method with different variants of the pioneer, called SPOT-ALIGN~\cite{duarte2022sign}, on How2Sign and PHOENIX-2014T. We also provide the results on CSL-Daily as a baseline.

Table~\ref{Tab:comparison_SOTA_how2sign} and Table~\ref{Tab:comparison_SOTA_PHOENIX} show the comparisons between our approach and SPOT-ALIGN~\cite{duarte2022sign} on How2Sign and PHOENIX-2014T, respectively. SPOT-ALIGN builds the final combination (COMB) model by integrating its primary cross-modal (CM) model with an auxiliary retrieval model (sign recognition (SR) model for How2Sign and Translation~\cite{camgoz2020sign} for PHOENIX-2014T). Our method outperforms the COMB model, which achieves best results in SPOT-ALIGN, by large margins, achieving +22.4 T2V and +28.0 V2T R@1 improvements on How2Sign, +13.7 T2V and +17.1 V2T R@1 improvements on PHOENIX-2014T. It is worth mentioning that the SPOT-ALIGN conducts three rounds of sign spotting~\cite{albanie2020bsl,momeni20_bsldict} and encoder training. In contrast, we simplify the training of sign encoder and only perform a single round of training on pseudo-labeled data. We also provide a baseline on CSL-Daily dataset as shown in Table~\ref{Tab:baseline_CSL}, demonstrating that our model can be generalized to various sign languages.

\subsection{Ablation Study \label{Subsec:Exp_ablation}}
We conduct all ablation studies on the most challenging How2Sign dataset.

\noindent\textbf{The Effectiveness of Each Proposed Component.}
Table~\ref{Tab:ablation_overall} shows the ablation of each component. We first build a baseline where the sign encoder is the domain-agnostic one and contrastive learning is trained with the standard contrastive loss~\cite{radford2021learning}. Then we add the proposed sign encoder (SE), cross-lingual contrastive learning (CLCL) and text augmentation (TA) to the baseline step by step. SE encodes domain-relevant and discriminative features, yielding +1.8 T2V and +4.5 V2T R@1 gains. The proposed CLCL significantly boosts the performance by +20.7 T2V and +19.1 V2T R@1 improvements, demonstrating that identifying fine-grained sign-to-word mappings is essential for sign language retrieval. The introduction of TA further promotes the retrieval task, achieving 56.6 R@1 on V2T and 51.6 R@1 on T2V, respectively.

\noindent\textbf{Various Sign Encoders.} As described in Section~\ref{Subsec:SFDA}, our sign encoder is composed of a domain-agnostic sign encoder (Single-Ag)~\cite{varol2021read} and a domain-aware sign encoder (Single-Aw). We first report the results of each individual encoder in Table~\ref{Tab:ablation_video_feature}. Next, we study the different ways to integrate the features extracted by these two encoders, including average and weighted sum as defined in Eq.~\ref{eq:feature_fusion}. The results are also shown in Table~\ref{Tab:ablation_video_feature}. We experimentally find that the weighted sum strategy yields best results.

\begin{table}[t]
 \scriptsize
 \centering
 \setlength{\tabcolsep}{4pt}
\begin{tabular}{l|ccc|ccc}
\toprule
\multirow{2}{*}{Method}& \multicolumn{3}{c|}{T2V} & \multicolumn{3}{c}{V2T} \\ 
  & R@1  & R@5  & R@10 & R@1  & R@5  & R@10\\
 \midrule
Baseline &31.5 & 49.6 & 57.9   & 26.4 & 44.4 & 52.8 \\
\midrule
+SE&33.3 & 51.9 & 59.8 & 30.9 & 48.8 & 56.4  \\
 % & \checkmark & &52.2 & 66.8 & 71.6 &  47.8 & 61.2 & 67.0  \\
  +SE+CLCL& 54.0 & 67.1 & 71.9 & 50.0 & 63.2 & 68.1  \\
 %& \checkmark & \checkmark & 53.1 & 68.0 & 73.4  & 47.3 & 62.9 & 67.0   \\
 +SE+CLCL+TA& \bf56.6 & \bf69.9 & \bf74.7  & \bf 51.6 & \bf64.8 & \bf70.1  \\
\bottomrule
\end{tabular}
\caption{Ablation study of each proposed component. SE: sign encoder; CLCL: cross-lingual contrastive learning; TA: text augmentation.}
\label{Tab:ablation_overall}
\vspace{-1em}
\end{table}

\begin{table}[t]
 \scriptsize
 \centering

 \setlength{\tabcolsep}{4pt}
\begin{tabular}{l|ccc|ccc}
\toprule
\multirow{2}{*}{Encoder} & \multicolumn{3}{c|}{T2V} & \multicolumn{3}{c}{V2T} \\
  &  R@1  & R@5  & R@10 & R@1  & R@5  & R@10\\
 \midrule
Single-Ag~\cite{varol2021read} &  53.1 & 68.0 & 73.4 &   47.3 & 62.9 & 67.0  \\
Single-Aw & 54.1 & 67.5 & 73.1   & 49.1 & 61.8 & 67.4  \\
\midrule
Fusion-Average &  54.7 & 68.7 & 73.8 &    49.6 & 63.7 & 69.0 \\
\bf{Fusion-Weighted Sum} &  \bf56.6 & \bf69.9 & \bf74.7  & \bf51.6 & \bf64.8 & \bf70.1  \\
\bottomrule
\end{tabular}
\caption{Results of domain-agnostic sign encoder (Single-Ag) and domain-aware sign encoder (Single-Aw). We also study different ways to integrate the features extracted by Single-Ag and Single-Aw, including average and weighted sum.} 
\label{Tab:ablation_video_feature}
\vspace{-5mm}
\end{table}

\noindent\textbf{Variants of Cross-Lingual Contrastive Learning.} 
As described in Section~\ref{Subsec:SWCL} and illustrated in Figure~\ref{fig:overview_B}, through the use of a combination of softmax, multiplication, sum, and average operations, we convert the fine-grained cross-lingual (sign-to-word) similarity to the coarse-grained sign-video-to-text similarity to enable contrastive learning. We refer to this process as ``Softmax''. Here we study two variants termed ``Mean'' and ``Max''. The ``Mean'' and ``Max' strategies simply replace the combination of softmax, multiplication and sum operations with a simple mean operation and a max operation, respectively. The results are shown in Table~\ref{Tab:ablation_sim_calcu}. We observe that the ``Mean'' strategy performs worst since it merely evaluates the overall similarity of a text and a sign video and ignores the fine-grained sign-to-word mappings during training. The ``Max'' strategy identifies hard sign-to-word mappings, \textit{i.e.}, one sign is associated with the most similar word, and vice versa. Nevertheless, since we do not have the ground-truth of sign-to-word mappings, it is challenging for models to identify the confident one-to-one mappings in a weakly supervised manner, as discussed in previous works on multiple instance learning~\cite{momeni20_bsldict}. In contrast, our default ``Softmax'' strategy localizes the soft sign-to-word mappings and achieves the best results.

\begin{table}[t]
 \scriptsize
 \centering
 \setlength{\tabcolsep}{4pt}
 
\begin{tabular}{l|ccc|ccc}
\toprule
\multirow{2}{*}{Strategy} & \multicolumn{3}{c|}{T2V} & \multicolumn{3}{c}{V2T} \\
 & R@1 & R@5 & R@10 & R@1 & R@5 & R@10 \\
 \midrule 

Mean & 33.1 & 52.5 & 59.7 & 29.8 & 47.8 & 55.3 \\
Max & 42.2 & 59.7 & 66.0 & 38.5 & 55.1 & 62.0 \\
\bf{Softmax}    & \bf56.6 & \bf69.9 & \bf74.7 & \bf51.6 & \bf64.8 & \bf70.1 \\
\bottomrule
\end{tabular}
\caption{Study on different strategies to identify the fine-grained sign-to-word mappings in cross-lingual contrastive learning.}
\label{Tab:ablation_sim_calcu}
     \vspace{-2mm}
\end{table}

\noindent\textbf{Study of Text Augmentations.}
In Table~\ref{Tab:ablation_text_aug}, we study three text augmentations described in Section~\ref{Subsec:TA}: random delete (RD), synonym replacement (SR) and random swap (RS). We observe a slight performance drop when introducing RD and SR into training. Sign language retrieval is not only a video-text retrieval task, but also a cross-lingual retrieval challenge, deletion and replacement may break the intrinsic sign-to-word mappings. In contrast, RS augmentation preserves the semantics of texts and we observe a +2.6 T2V and a +1.6 V2T R@1 improvements over the counterpart without any text augmentations.

\begin{table}[t]
 \centering
 \scriptsize
 \setlength{\tabcolsep}{4pt}
\begin{tabular}{c|ccc|ccc}
\toprule
\multirow{2}{*}{\makecell[c]{Text\\Augmentation}} 
&\multicolumn{3}{c|}{T2V} & \multicolumn{3}{c}{V2T} \\
&  R@1  & R@5  & R@10 & R@1  & R@5  & R@10\\
 \midrule
None & 54.0 & 67.1 & 71.9 &   50.0 & 63.2 & 68.1   \\
RD & 52.7 & 66.3 & 71.6   & 47.6 & 62.1 & 67.4   \\
SR &53.9 & 68.7 & 73.0  & 47.5 & 61.2 & 65.9  \\
\bf{RS} & \bf56.6 & \bf69.9 & \bf74.7 &   \bf51.6 & \bf64.8 & \bf70.1 \\
\bottomrule
\end{tabular}
\caption{Ablation study on different text augmentations. RD: random delete; SR: synonym replacement; RS: random swap.}
\label{Tab:ablation_text_aug}
\vspace{-2mm}
\end{table}

\noindent\textbf{The Effects of CLIP Initialization.} In our framework, the vision and text Transformer are initialized with the CLIP's vision encoder and text encoder. Note that the input modality of our vision Transformer and that of CLIP's vision encoder are different. Ours takes a feature sequence as input while the input of CLIP's image encoder is a set of image patches. In Table~\ref{Tab:ablation_vision_encoder}, we compare a randomly initialized baseline with the one initialized by CLIP. We find that CLIP-initialization significantly improves the performance, yielding +11.4 T2V and +11.0 V2T R@1 improvements though the input originates from a completely distinct modality.

\noindent\textbf{Study on Text Transformer.}
The text Transformer in our model is initialized with the CLIP's text encoder.
Considering the generalization capability of the pre-trained CLIP, we attempt to freeze the shallow layers of our text transformer. As shown in Table~\ref{Tab:ablation_text_encoder}, the performance gradually decreases as the number of the frozen layers increases. CLIP is trained on large-scale image-text pairs, however, the domain gap between image-text data and sign-video-text data is non-negligible. Though CLIP shows promising transfer capacity, it is optimal to fine-tune the whole model on target datasets.

\begin{table}[t!]
 \scriptsize
 \centering
 \setlength{\tabcolsep}{4pt}
 
\begin{tabular}{c|ccc|ccc}

\toprule
\multirow{2}{*}{\makecell[c]{ Initialization}}  & \multicolumn{3}{c|}{T2V} & \multicolumn{3}{c}{V2T} \\
 &R@1  & R@5  & R@10 & R@1  & R@5  & R@10\\
 \midrule
Random & 45.2&	60.3&	67.5& 40.6&	54.5&	59.7\\
\textbf{CLIP} & \bf56.6&	\bf69.9& \bf74.7&  \bf51.6&	\bf64.8&	\bf70.1\\

\bottomrule
\end{tabular}
\caption{Comparison between CLIP initialization and random initialization.}
\label{Tab:ablation_vision_encoder}
\vspace{-3mm}
\end{table}

\begin{table}[t]
 \scriptsize
 \centering
 \setlength{\tabcolsep}{4pt}

\begin{tabular}{c|ccc|ccc}

\toprule
\multirow{2}{*}{\makecell[c]{Frozen  layers}} & \multicolumn{3}{c|}{T2V} & \multicolumn{3}{c}{V2T} \\
  &R@1  & R@5  & R@10 & R@1  & R@5  & R@10\\
 \midrule

 \textbf{None}   & \bf56.6 & \bf69.9 & \bf74.7   & \bf51.6 & \bf64.8 & \bf70.1 \\
 
\midrule

1   & 54.4 & 68.2 & 73.4   & 49.7 & 63.5 & 68.8   \\
2   & 53.8 & 68.6 & 73.3   & 48.5 & 62.8 & 67.9   \\ 
6  & 51.7 & 66.6 & 71.1   & 46.2 & 60.9 & 66.6   \\
12   & 49.5 & 65.2 & 71.4   & 44.7 & 59.7 & 65.2   \\

\bottomrule
\end{tabular}
\caption{Study of freezing different layers of text Transformer.}
\label{Tab:ablation_text_encoder}
\vspace{-2mm}
\end{table}

\section{Conclusion \label{Sec:Conclusion}}
In this paper, we propose a novel framework named \textbf{C}ross-l\textbf{i}ngual \textbf{Co}ntrastive learning (CiCo) for recently introduced sign language retrieval task. We formulate sign language retrieval as a cross-lingual retrieval task as well as a video-text retrieval problem. CiCo models the fine-grained cross-lingual mappings between sign videos and texts via the proposed cross-lingual contrastive learning. We also introduce a sign encoder, which is composed of a domain-agnostic encoder and a domain-aware one, to extract discriminative and domain-aligned features. Our Cico outperforms the pioneer SPOT-ALIGN by large margins on How2Sign and PHOENIX-2014T benchmarks. We also provide a baseline on CSL-Daily. We hope our approach could serve as a solid baseline for future research.

\noindent \textbf{Acknowledgement.} This work was partially supported by National Natural Science Foundation of China (No. 62072112), National Key R\&D Program of China (No. 2020AAA0108301), and Scientific and Technological Innovation Action Plan of Shanghai Science and Technology Committee (No. 22511101502, 22511102202).

{\small
\bibliographystyle{ieee_fullname}
\bibliography{references}
}

\appendix
\section{More Experiments}

\renewcommand\arraystretch{1.0}
\noindent\textbf{CiCo vs CLIP.}\hspace{3pt} 
 We compare our approach CiCo with CLIP~\cite{radford2021learning}, which is one of the most representative vision-language models. CLIP can be easily generalized to sign language retrieval by replacing our cross-lingual contrastive learning with CLIP. The other settings including sign encoder and text augmentation still remain unchanged. As shown in Table~\ref{Tab:ablation_cico_vs_clip}, CiCo surpasses CLIP by +21.2 T2V and +20.9 V2T R@1 scores. The reason is that CLIP contrasts the overall features of two modalities, while our cross-lingual contrastive learning concentrates on identifying the fine-grained sign-to-word mappings during modeling global similarities of texts and sign videos. 

\begin{table}[t]
 \small
 \setlength{\tabcolsep}{4pt}
 \centering
\begin{tabular}{c|ccc|ccc}
\toprule
\multirow{2}{*}{Model} & \multicolumn{3}{c|}{T2V} & \multicolumn{3}{c}{V2T} \\ 
 &   R@1  & R@5  & R@10 & R@1  & R@5  & R@10\\
 \midrule

CLIP~\cite{radford2021learning} & 35.4&53.4&60.9&30.7&49.1&57.1  \\
\textbf{CiCo}  &\bf56.6 & \bf69.9 & \bf74.7 & \bf51.6 & \bf64.8 & \bf70.1 \\
\bottomrule
\end{tabular}
\caption{Comparison between Cico and CLIP.}
\label{Tab:ablation_cico_vs_clip}
\end{table}

\noindent\textbf{Different Strategies of Global Similarity Calculation in Cross-Lingual Contrastive Learning.}\hspace{3pt} As described in Section 3.3 and illustrated in Figure 3b, we adopt ``Mean'' strategy which averages sign-to-text similarities and word-to-video similarities to obtain the global video-to-text similarity and text-to-video similarity, respectively. In Section 4.3 of the main paper, we study different strategies to identify the fine-grained sign-to-word mappings, now we investigate different ways of global similarity calculation.  Table~\ref{Tab:ablation_global_calcu} shows the results of two variants termed ``Max'' and ``Softmax'' besides the default ``Mean'' strategy. ``Max'' assigns global similarity with the maximum score of sign-to-text similarities (or word-to-video similarities). ``Softmax'' stands for a combination of Softmax, multiplication and sum (refer to Section 4.3 for details). The default ``Mean'' strategy achieves the best result.

\noindent\textbf{Sliding Window Stride in Sign Encoder.}\hspace{3pt} Our sign encoder adopts a sliding window manner to extract features of continuous sign videos. The default sliding window stride is set as 1. We vary the stride and show the results in Table~\ref{Tab:ablation_stride}. Setting stride as 1 yields the best performance.

\begin{table}[t]
 \small
 \centering
 \setlength{\tabcolsep}{4pt}
\begin{tabular}{l|ccc|ccc}
\toprule
\multirow{2}{*}{Strategy} & \multicolumn{3}{c|}{T2V} & \multicolumn{3}{c}{V2T} \\
 & R@1 & R@5 & R@10 & R@1 & R@5 & R@10 \\
\midrule

Max     & 21.1 & 38.0 & 46.4 & 17.8 & 34.9 & 42.9 \\

Softmax & 32.6 & 50.3 & 58.2 & 29.0 & 46.6 & 54.0 \\
\bf{Mean}    & \bf56.6 & \bf69.9 & \bf74.7 & \bf51.6 & \bf64.8 & \bf70.1 \\
\bottomrule
\end{tabular}
\caption{Study on different strategies of global similarity calculation in cross-lingual contrastive learning.}
\label{Tab:ablation_global_calcu}
\end{table}

\begin{table}[t]
 \small
 \centering
 \setlength{\tabcolsep}{4pt}
\begin{tabular}{c|ccc|ccc}
\toprule
\multirow{2}{*}{Stride} & \multicolumn{3}{c|}{T2V} & \multicolumn{3}{c}{V2T} \\
 & R@1 & R@5 & R@10 & R@1 & R@5 & R@10 \\
\midrule
\bf{1} & \bf56.6 & \bf69.9 & \bf74.7 & \bf51.6 & \bf64.8 & \bf70.1 \\
						
2   &44.8&60.5&68.1&39.7&55.5&63.0 \\
4   &24.3&42.3&49.8&14.4&30.2&37.4 \\
8 &23.6&	40.8&	49.1&	15.3	&31.5	&39.6\\
\bottomrule
\end{tabular}
\caption{Study on different sliding window strides used in sign encoder.}
\label{Tab:ablation_stride}
\end{table}

\begin{figure*}[t]
	\centering
	\scriptsize
 
		\begin{tabular}{ccc}
			\subfloat{\includegraphics[width=0.3\linewidth]{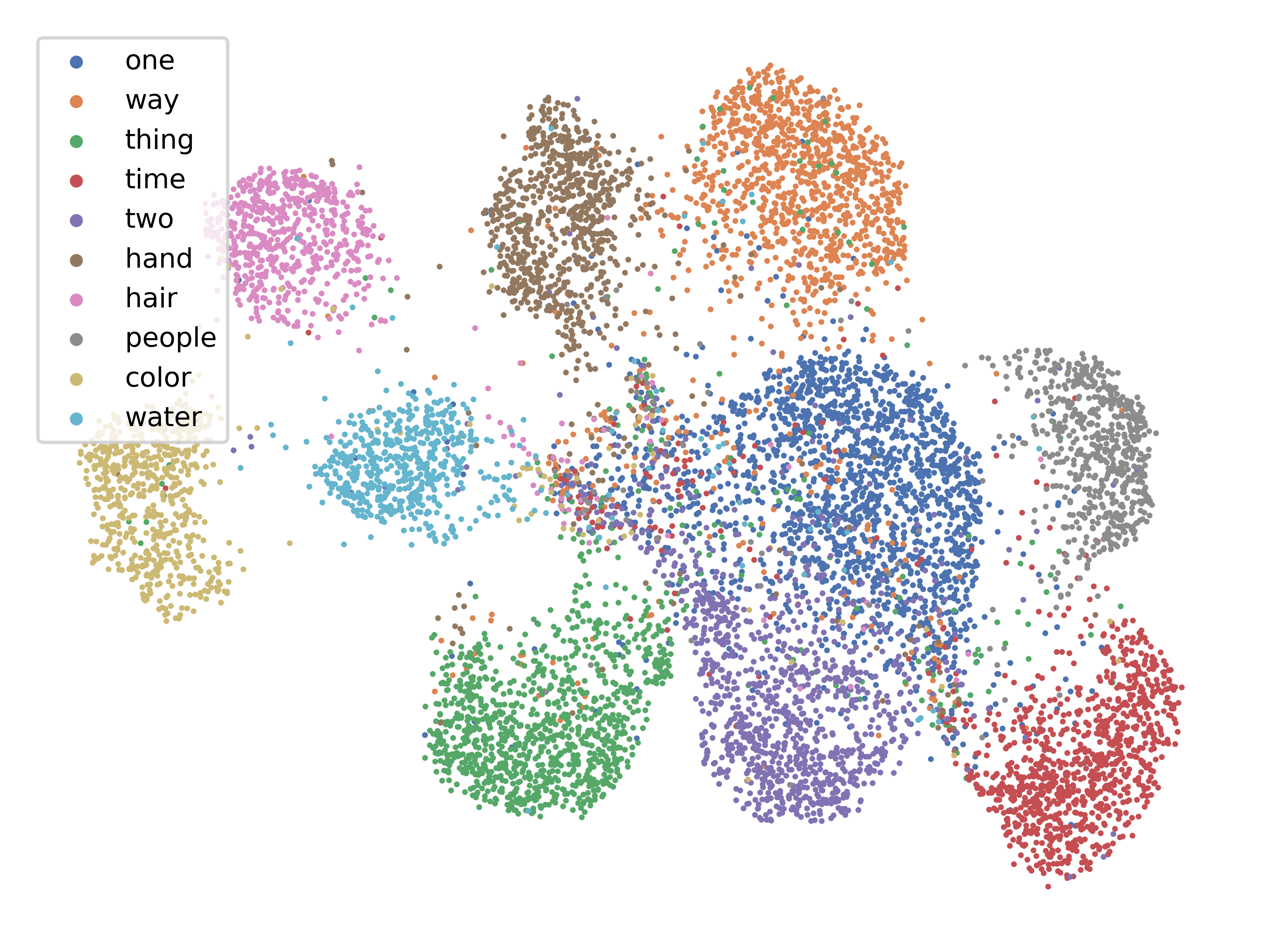}} & 
			\subfloat{\includegraphics[width=0.3\linewidth]{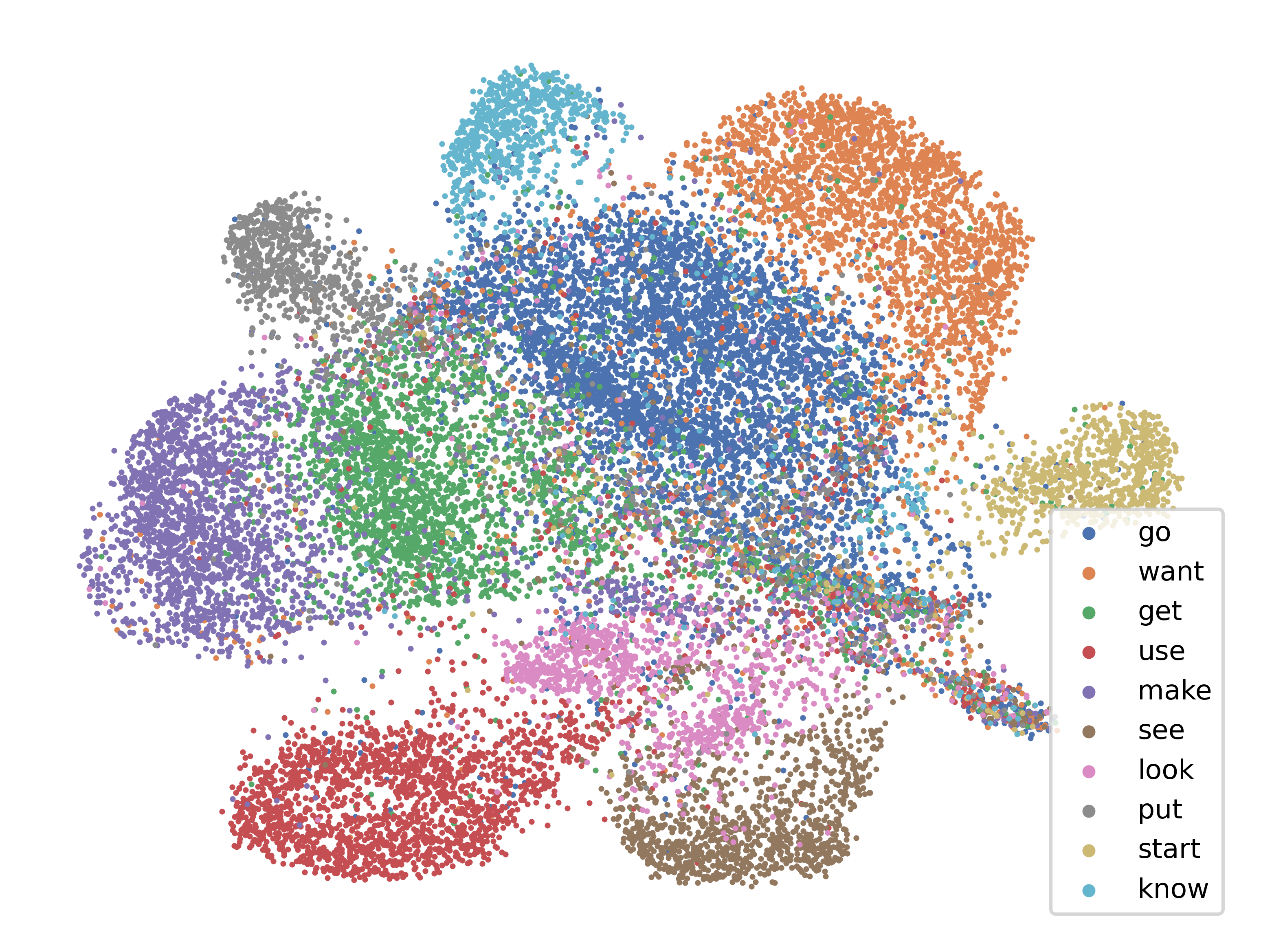}} & 
			\subfloat{\includegraphics[width=0.3\linewidth]{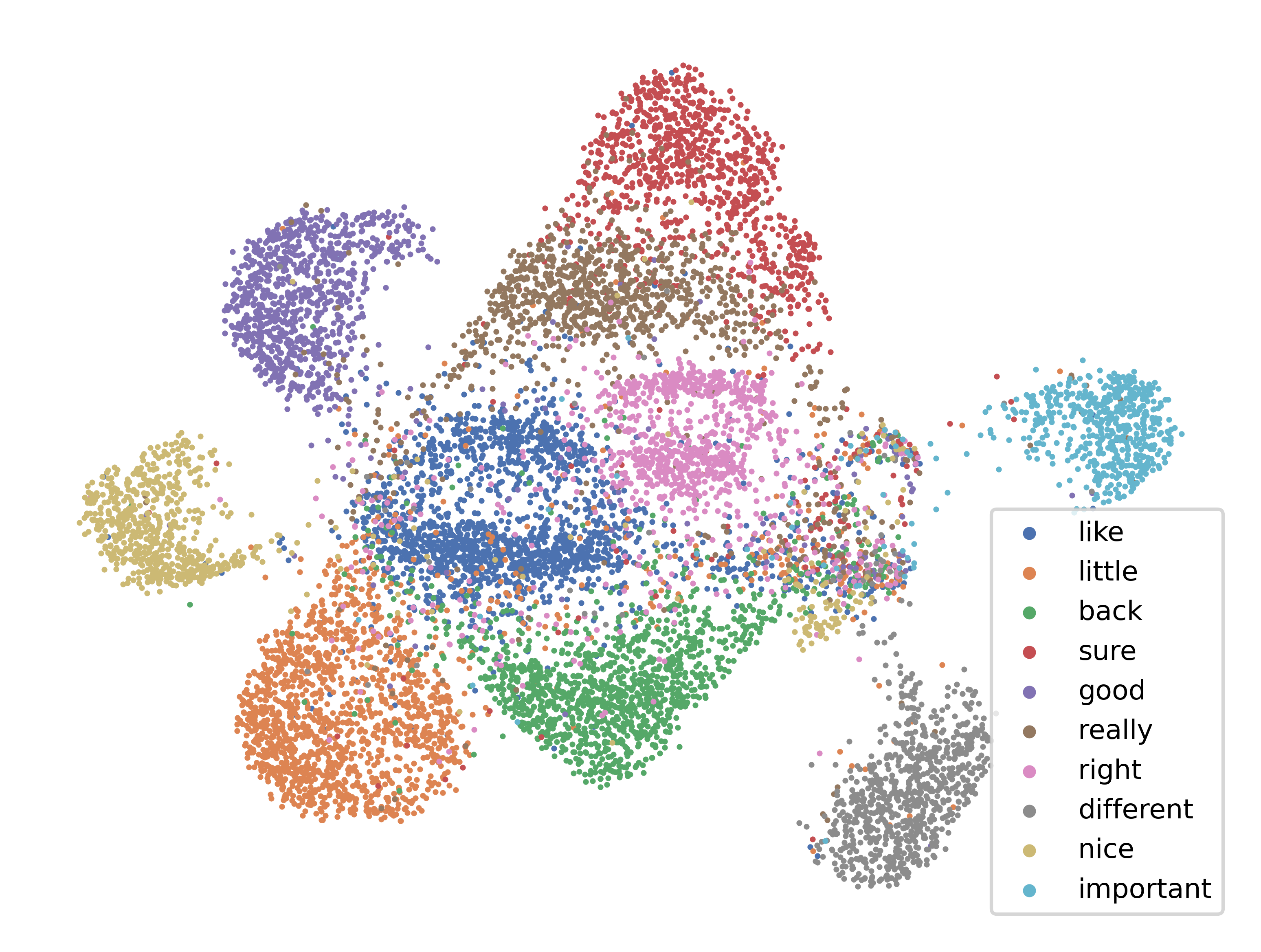}} 

			\\
			
			(a) Top-10 Nouns. & (b) Top-10 Verbs. & (c) Top-10 Adjective/Adverbs. 
		\end{tabular}
	\caption{Feature visualization of sign video clips. We map features extracted by our sign encoder to 2D space with UMAP~\cite{mcinnes2018umap}.}

 \label{fig:vis_UMAP}
\end{figure*}

\begin{figure}[b]
     \centering
     \begin{subfigure}[t]{0.7\linewidth}
         \includegraphics[width=\textwidth]{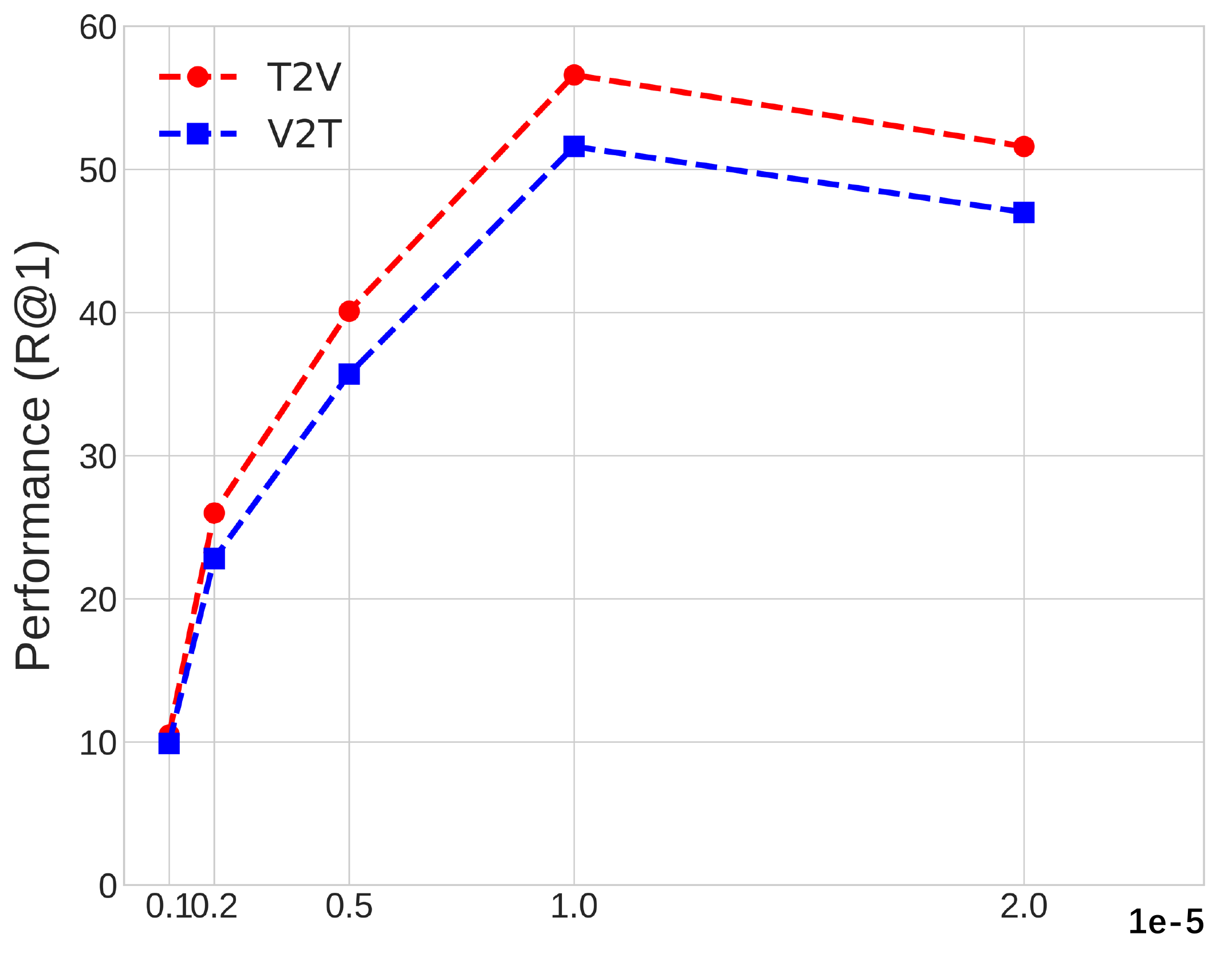}
         \caption{Learning rate.}
             \label{Fig:ablation_lr}
      \end{subfigure}
        \hspace{0.1cm}
     \begin{subfigure}[t]{0.7\linewidth}
         \includegraphics[width=\textwidth]{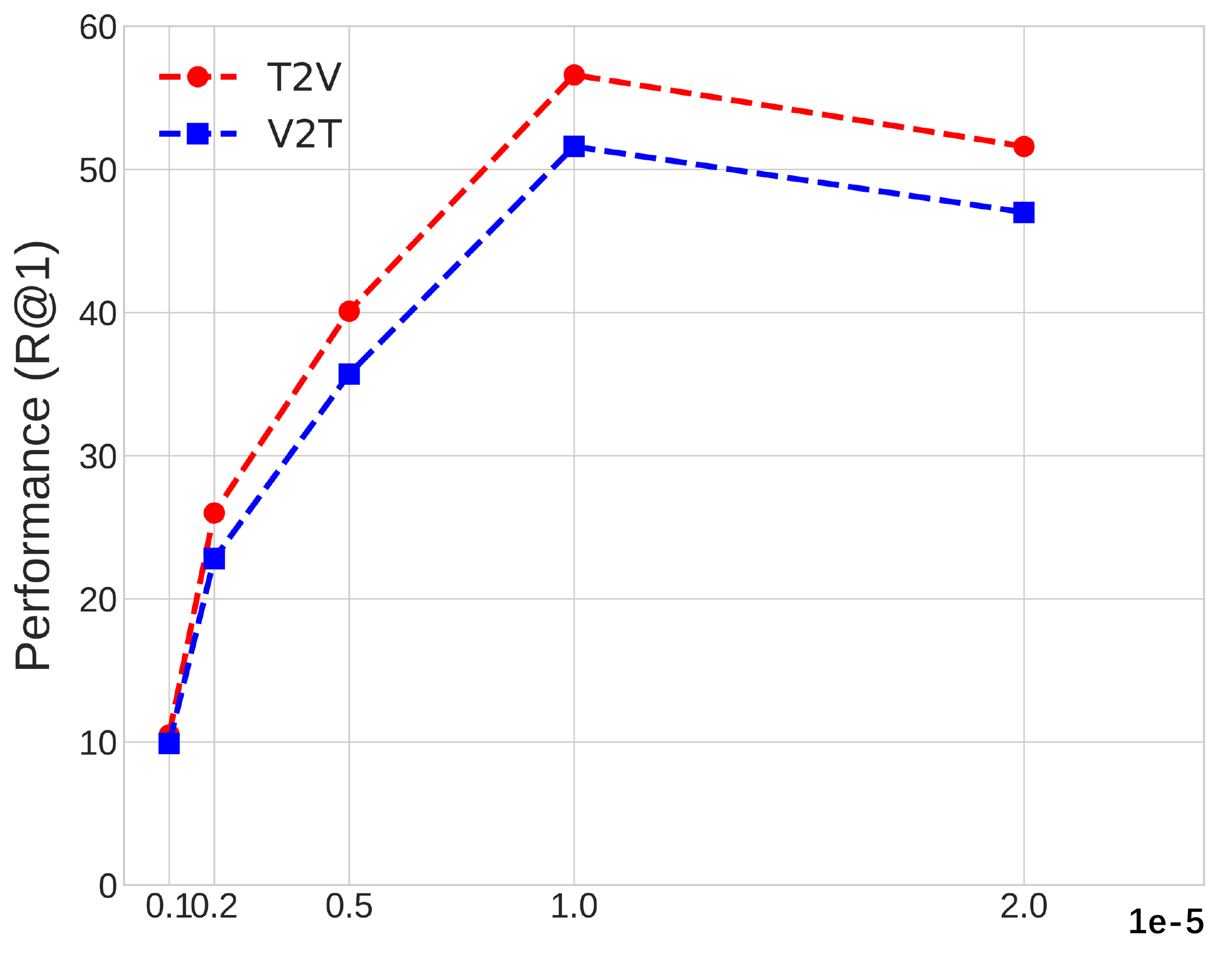}
        \caption{Batch size.}
        \label{Fig:ablation_bz}

     \end{subfigure}
    \caption{Study on fine-tinning hyper-parameters in contrastive learning.}
    \label{Fig:ablation:learning_strategy}
\end{figure}

\noindent\textbf{Fine-Tuning Hyper-Parameters.} \hspace{3pt} Recall that in the training of cross-lingual contrastive learning, our vision transformer and text transformer are initialized by the image encoder and text encoder in CLIP (ViT-B/32)~\cite{radford2021learning}. Here we study the fine-tuning hyper-parameters, \textit{i.e.}, learning rate in Figure~\ref{Fig:ablation_lr}  and batch size in Figure~\ref{Fig:ablation_bz}. A learning rate of 1e-5 yields best result. The increase of batch size sustainably promotes the performance. In our experiment, we set the batch size to 512 due to the limited GPU memory.

\noindent\textbf{Other Hyper-Parameters.}\hspace{3pt} There are four remaining hyper-parameters in CiCo: 1) $\alpha$ defined in Eq.(1) controls the weights of features extracted by domain-agnostic sign encoder and domain-aware sign encoder; 2) $\beta$ defined in Eq.(3) controls the weights of sign-video-to-text contrast and text-to-sign-video contrast; 3) the temperature $\sigma$ of row-wise and column-wise Softmax; 4) the maximum length of sign clip feature $L$. The studies are shown in Table~\ref{Tab:ablation_alpha}, Table~\ref{Tab:ablation_beta}, Table~\ref{Tab:ablation_sigma} and Table~\ref{Tab:ablation_L}, respectively.

\renewcommand\arraystretch{1.0}

\begin{table}[t]
  \small

 \centering
 \setlength{\tabcolsep}{4pt}
\begin{tabular}{c|ccc|ccc}

\toprule
\multirow{2}{*}{$\alpha$} & \multicolumn{3}{c|}{T2V} & \multicolumn{3}{c}{V2T} \\
  & R@1  & R@5  & R@10 & R@1  & R@5  & R@10\\
 \midrule
0.2 & 53.0 & 67.5 & 72.5  & 47.6 & 62.7 & 67.2 \\
0.4 & 55.4 & 68.7 & 74.0 & 49.9 & 62.5 & 68.6  \\
0.6 & 55.1 & 68.5 & 73.4  & 49.6 & 63.9 & 68.9   \\
\bf{0.8} & \bf56.6 & \bf69.9 & \bf74.7 & \bf51.6 & \bf64.8 & \bf70.1 \\
\bottomrule
\end{tabular}
\caption{Study of $\alpha$ defined in Eq.(1).}
\label{Tab:ablation_alpha}
\end{table}

\begin{table}[t]
  \small

 \centering
 \setlength{\tabcolsep}{4pt}
\begin{tabular}{c|ccc|ccc}
\toprule
\multirow{2}{*}{$\beta$} &\multicolumn{3}{c|}{T2V} & \multicolumn{3}{c}{V2T} \\
  &R@1  & R@5  & R@10 & R@1  & R@5  & R@10\\
 \midrule
0.0 & 39.0 & 56.3 & 63.1 & 26.6 & 49.9 & 57.8 \\
0.2 & 44.8 & 62.1 & 68.1 & 39.8 & 55.5 & 62.5 \\
0.4 & 45.8 & 62.4 & 68.7 & 40.6 & 57.7 & 64.1 \\
\textbf{0.5}  &\bf56.6 & \bf69.9 & \bf74.7  & \bf51.6 & \bf64.8 & \bf70.1 \\
0.6 & 54.9 & 69.6 & 74.5 & 49.6 & 63.5 & 68.6 \\
0.8 & 54.1 & 68.7 & 73.3 & 48.3 & 62.1 & 67.8 \\
1.0 &  52.5 & 67.1 & 72.1 &    48.8 & 62.8 & 67.4  \\
 
\bottomrule
\end{tabular}
\caption{Study of $\beta$ defined in Eq.(3).}
\label{Tab:ablation_beta}
\end{table}

\begin{table}[t]
  \small

 \centering
 \setlength{\tabcolsep}{4pt}
\begin{tabular}{c|ccc|ccc}
\toprule
\multirow{2}{*}{$\sigma$} & \multicolumn{3}{c|}{T2V} & \multicolumn{3}{c}{V2T} \\
  & R@1  & R@5  & R@10 & R@1  & R@5  & R@10\\
 \midrule
7e-04 & 41.3 & 58.9 & 65.5  & 38.2 & 54.4 & 61.3  \\
7e-03 & 42.6 & 59.6 & 65.5  & 39.5 & 54.7 & 61.9  \\
\bf{7e-02} & \bf56.6 & \bf69.9 & \bf74.7 & \bf51.6 & \bf64.8 & \bf70.1 \\
7e-01 & 31.9 & 49.9 & 57.8 & 28.6 & 45.8 & 53.9 \\
%7 & 32.2 & 51.5 & 58.9  & 28.3 & 45.7 & 53.8  \\
\bottomrule
\end{tabular}
\caption{Study of the temperature $\sigma$ used in row-wise and column-wise Softmax.}
\label{Tab:ablation_sigma}
\end{table}

\begin{table}[t]
  \small

 \centering
 \setlength{\tabcolsep}{4pt}
\begin{tabular}{c|ccc|ccc}

\toprule
\multirow{2}{*}{$L$} & \multicolumn{3}{c|}{T2V} & \multicolumn{3}{c}{V2T} \\
  & R@1  & R@5  & R@10 & R@1  & R@5  & R@10\\
 \midrule
4 & 17.3 & 31.2 & 38.6 & 14.3 & 26.8 & 34.1  \\
8 & 38.2 & 55.4 & 62.3 & 34.1 & 50.2 & 56.4  \\
16 & 50.9 & 66.6 & 72.0 & 45.9 & 60.3 & 66.7  \\
32 & 53.6 & 67.3 & 73.5 & 48.9 & 61.7 & 67.8  \\
\bf{64} & \bf{56.6} & \bf{69.9} & \bf{74.7} & \bf{51.6} & \bf{64.8} & \bf{70.1}  \\ 
\bottomrule
\end{tabular}
\caption{Study of maximum length of sign clip feature $L$.}
\label{Tab:ablation_L}
\end{table}

\section{Qualitative Results}
{\noindent \textbf{Visualization of the Identified Sign-to-Word Mappings.}}\hspace{3pt}
Recall that in cross-lingual contrastive learning, we implicitly identify the sign-to-word mappings by calculating the fine-grained cross-lingual similarities (see Figure~3b of the main paper). Once the model is well optimized, we could infer the input texts and sign videos to produce a cross-lingual similarity matrix, which approximately reflects the sign-to-word mappings. For each word, we could identify its corresponding sign which has the maximal activation value. After that, the sign-to-word mapping is established. In Figure~\ref{fig:vis_UMAP}, we utilize UMAP~\cite{mcinnes2018umap} to visualize the features of the identified sign video clips for top-10 nouns, verbs and adjectives/adverbs within the How2Sign~\cite{Duarte_CVPR2021} vocabulary. The features of sign video clips associated with the same word form a compact cluster, demonstrating that our approach could identify the sign-to-word mappings during training.

\begin{figure*}[b]
     \centering
     \begin{subfigure}[b]{\linewidth}
         \centering
         \includegraphics[width=\textwidth]{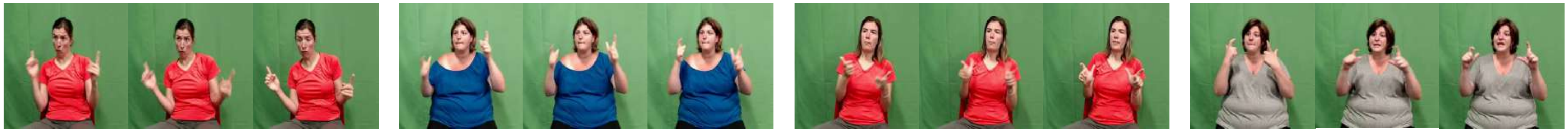}
         \caption{``Big''.}
      \end{subfigure}
     \hfill
     
     \centering
     \begin{subfigure}[b]{\linewidth}
         \centering
         \includegraphics[width=\textwidth]{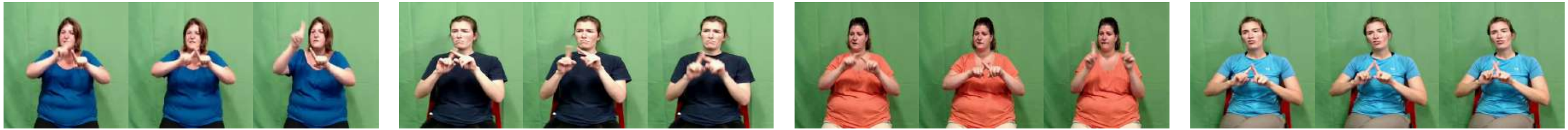}
         \caption{``Different''.}

     \end{subfigure}
     \hfill

     \centering
     \begin{subfigure}[b]{\linewidth}
         \centering
         \includegraphics[width=\textwidth]{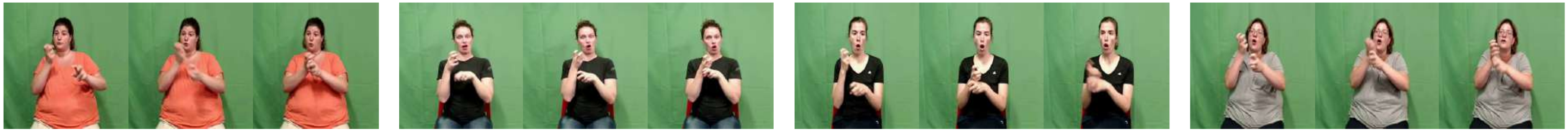}
         \caption{``Hard''.}

     \end{subfigure}
     \hfill
     
     \centering
     \begin{subfigure}[b]{\linewidth}
         \centering
         \includegraphics[width=\textwidth]{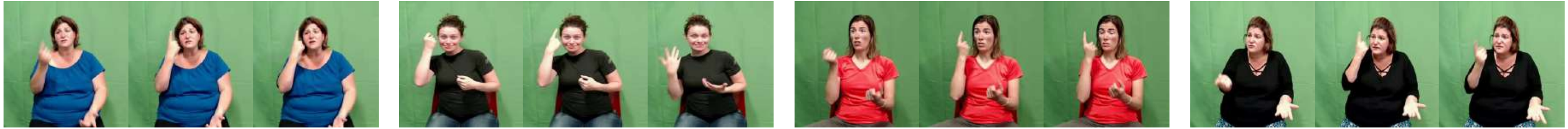}
         \caption{``Understand''.}

     \end{subfigure}
     \hfill

     \centering
     \begin{subfigure}[b]{\linewidth}
         \centering
         \includegraphics[width=\textwidth]{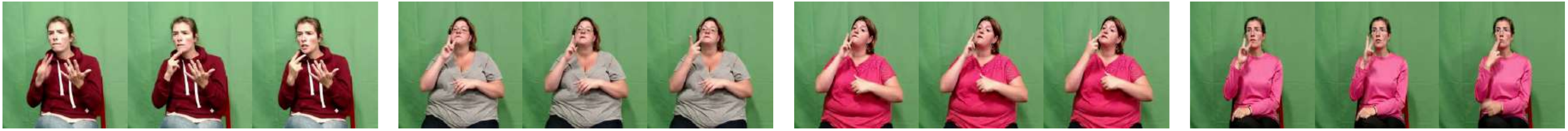}
         \caption{``Vegetable''.}

     \end{subfigure}
     \hfill
     
     \centering
     \begin{subfigure}[b]{\linewidth}
         \centering
         \includegraphics[width=\textwidth]{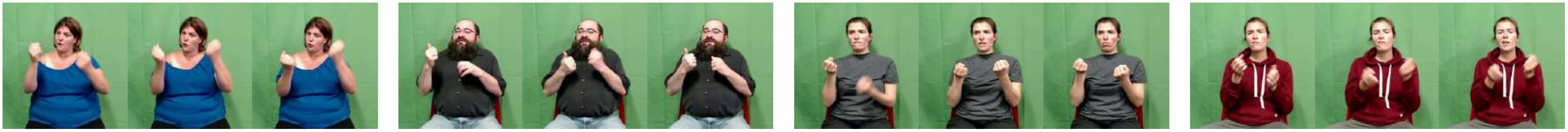}
         \caption{``Vehicle''.}

     \end{subfigure}
     \hfill

     \centering
     \begin{subfigure}[b]{\linewidth}
         \centering
         \includegraphics[width=\textwidth]{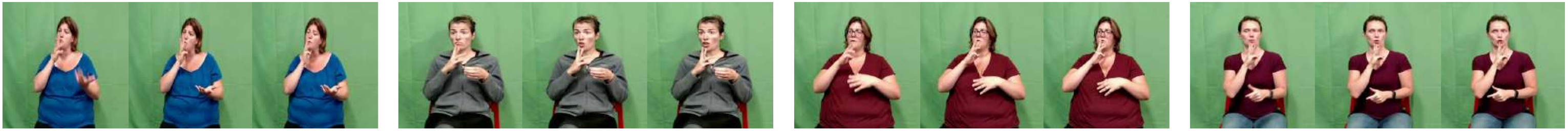}
         \caption{``Water''.}

     \end{subfigure}
     \hfill
     
     \centering
     \begin{subfigure}[b]{\linewidth}
         \centering
         \includegraphics[width=\textwidth]{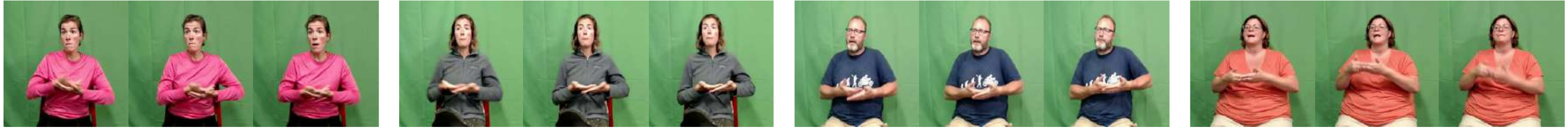}
         \caption{``Baby''.}

     \end{subfigure}
     \hfill

     \caption{More examples of cross-lingual (sign-to-word) mappings identified by our approach on How2Sign~\cite{Duarte_CVPR2021} dataset.}
     \label{fig:more_mappings}
    
\end{figure*}

{\noindent \textbf{More Examples of Sign-to-Word Mappings.}}\hspace{3pt} We visualize a collection of signs associated with the words \{``Big'', ``Different'', ``Hard'', ``Understand'', ``Vegetable'', ``Vehicle'', ``Water'', ``Baby''\} in Figure~\ref{fig:more_mappings}. The mappings are automatically identified by our CiCo.

\end{document}